\documentclass[letterpaper, 10 pt, conference]{ieeeconf}

\IEEEoverridecommandlockouts

\usepackage[utf8]{inputenc}

\usepackage{amsmath}
\usepackage{hyperref}
\usepackage[nolist,nohyperlinks]{acronym}
\usepackage{nomencl}
\makenomenclature
\usepackage{tikz}       % graphic symbols
\usepackage{etoolbox}   % ifstrequal
\usepackage{amsfonts}   % math fonts
\usepackage[caption=false]{subfig}
\usepackage{bm}         % bold math
\usepackage{multirow}   % multi-row
\usepackage{array}      % vertical alignment
\usepackage{paralist} % enumerate and itemize within paragraphs

\acrodef{hsi}[HSI]{hue saturation color}
\acrodef{pca}[PCA]{principle component analysis}
\acrodef{ransac}[RANSAC]{random sample consensus}
\acrodef{ir}[IR]{infra red}
\acrodef{cht}[CHT]{circle Hough Transform}

\definecolor{junccolor}{RGB}{100,0,200}
\definecolor{tailcolor}{RGB}{200,0,0}
\definecolor{pendcolor}{RGB}{0,150,30}
\definecolor{eecolor}{RGB}{255, 150, 0}
\definecolor{graspcolor}{RGB}{200, 0, 150}

\newcommand{\baseline}{
	\the\dimexpr\fontdimen22\textfont2\relax
}

\newcommand{\R}{\mathbb{R}}

\graphicspath{{fig/}}

\usepackage{balance}
\usepackage{hyperref}

\title{\LARGE \bf
Geometry-Based Grasping of Vine Tomatoes}

\author{Taeke de Haan$^{1}$ \and Padmaja Kulkarni$^{1}$ \and Robert Babu\v{s}ka$^{1}$% <-this % stops a space
\thanks{$^{1}$Authors are with Department of Cognitive Robotics, Delft University of Technology, Delft, The Netherlands.
        {\tt\small taeke.dehaan@live.nl, (P.V.kulkarni, R.Babuska)@tudelft.nl}}%
}

\date{\today}

\begin{document}

% Title page must contain
% - paper title
% - each author's:
%   o name
%   o affiliation
%   o full address (mailing address, email address, and fax number), with the corresponding author clearly indicated,
% - abstract (no more than 200 words)
% - keywords (index terms)
% - beginning of the main text of the paper

\maketitle
\setcounter{footnote}{1}
%%%%%%%%%%%%%%%%%%%%%%%%%
%%%%%%% ABSTRACT %%%%%%%%
%%%%%%%%%%%%%%%%%%%%%%%%%
% TODO maximum of 200 word
\begin{abstract}
We propose a geometry-based grasping method for vine tomatoes. It relies on a computer-vision pipeline to identify the required geometric features of the tomatoes and of the  truss stem. The grasping method then uses a geometric model of the robotic hand and the truss to determine a suitable grasping location on the stem. This approach allows for grasping tomato trusses without requiring delicate contact sensors or complex mechanistic models and under minimal risk of da\-maging the tomatoes. Lab experiments were conducted to validate the proposed methods, using an RGB-D camera and a low-cost robotic manipulator. The success rate was 83\% to 92\%, depending on the type of truss. The codebase\footnote{\url{https://github.com/padmaja-kulkarni/taeke\_msc}} and the videos of the experiments\footnote{\url{https://youtu.be/SEW5gtK6E0I}} are available online. 
\end{abstract}
% TODO: add keywords

%%%%%%%%%%%%%%%%%%%%%%%%%
%%%%% INTRODUCTION %%%%%%
%%%%%%%%%%%%%%%%%%%%%%%%%
\section{Introduction}

Mechanization and automation have tremendously increased crop production over the course of history. Further automation of harvesting and food processing will diminish the impact of labor shortage, reduce the financial risk for farmers and increase agricultural efficiency \cite{bergerman2016robotics, foley2011solutions}. One of the main challenges is the handling of fragile and deformable organic objects without damaging them \cite{rodriguez2013grasping}. Tomato is an example of such a product. According to the Food and Agriculture Organization of the United Nations Statistics Database (FAOSTAT),\footnote{Public database \url{http://www.fao.org/faostat/en/\#compare}} tomato is the second most-produced vegetable crop, after potato. This paper focuses on vine tomatoes, also called truss or clustered tomatoes, which are marketing terms for tomatoes kept attached to the fruiting stem after the harvest. Currently, harvesting, weighing, and packaging of vine tomatoes rely solely on manual work.

% A solution is desired which may handle these fragile deformable objects while solely relying on geometric information and position-based control.

When applying robotic technology, one can prevent damaging deformable objects by using touch and force sensors or precise mechanistic models of the objects~\cite{lin2014picking},~\cite{zaidi2017model}. However, sensors are fragile, expensive, and complicate the cleaning and disinfection of the gripper. Obtaining and simulating mechanistic models for deformable and fragile objects is challenging and no readily applicable methods are available~\cite{sanchez2018robotic}.

\begin{figure}[htbp]
        \centering
        \includegraphics[width = 0.8\linewidth]{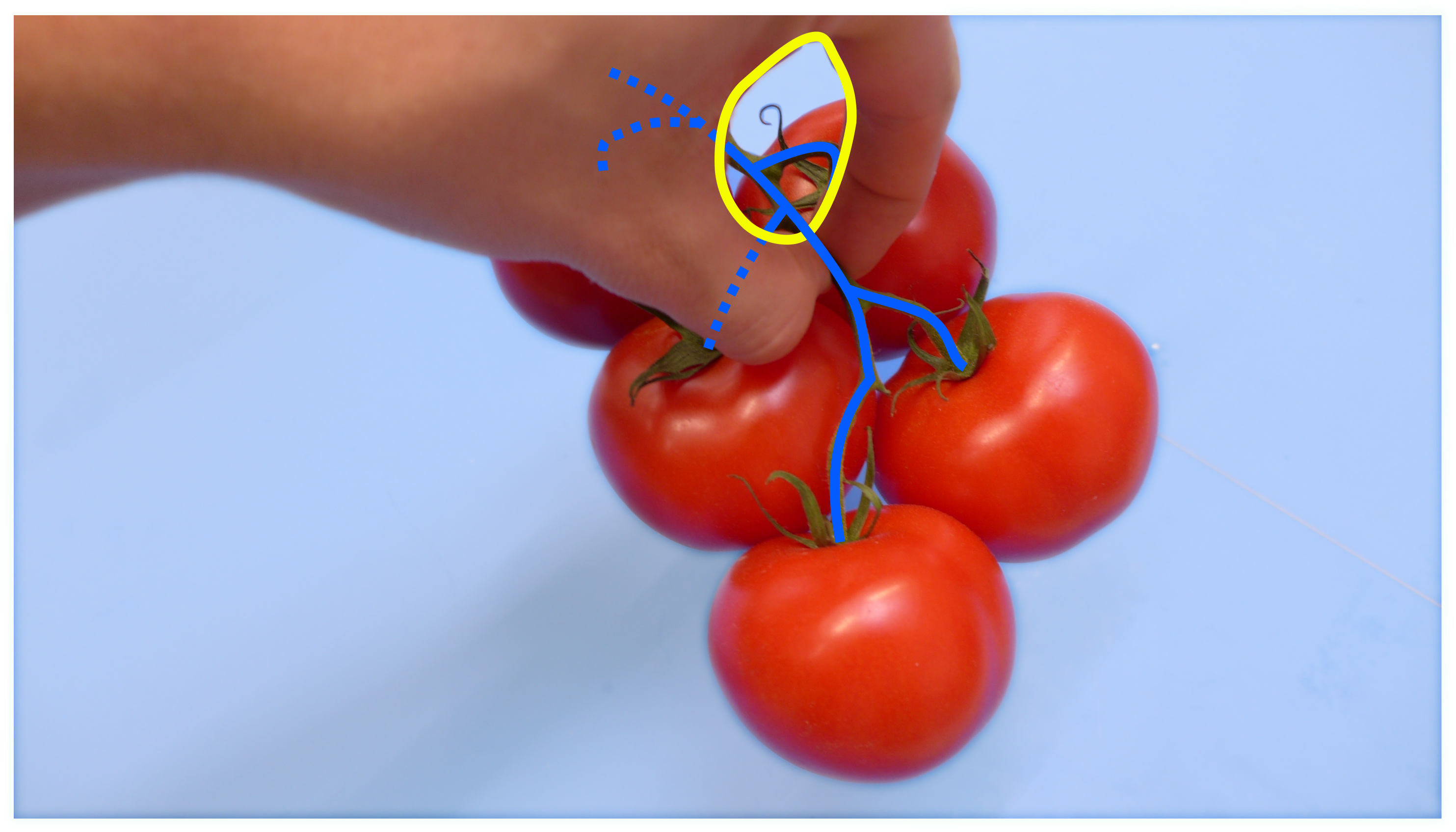}
        \caption{A human hand applies caging to a tomato truss. In this way, one can grasp the truss without applying force to the tomatoes~\cite{varava2016caging}.}
        % The ends of the blue curve are spread sufficiently far to prevent the truss from moving arbitrarily far from the yellow loop formed by the human hand.
        \label{fig:02:cage}
\end{figure}

This paper aims to overcome the above limitations by developing a novel geometry-based grasping method for vine tomatoes. We use caging as a grasp method that bounds the target without requiring tight contact between the object and end effector~\cite{kuperberg1990problems}. The general benefits of caging are low mechanical stress applied to the target object, robustness to minor control errors, and no need for force feedback or a mechanistic object model. We apply caging to the stem, using the specific features of its geometry, as illustrated in Fig.~\ref{fig:02:cage} and detailed in the sequel.

%These features, as defined by Varava et al.~\cite{varava2016caging}, can be described by a curve whose ends spread sufficiently far to prevent the loop formed by the end effector from moving arbitrarily far away from the object. In the case of vine tomato, pedicels are connected at multiple junctions to the peduncle. The end effector encloses the branches encapsulated by these junctions.

% The proposed method utilizes a geometric model of the robotic hand and truss to determine an optimal grasping location on the peduncle.

This work's main contribution is the development of a vision pipeline to identify a suitable grasping location on the peduncle of vine tomatoes and the subsequent geometry-based grasping approach. We validate the methods by means of lab experiments performed with a low-cost manipulator and an RGB-D camera.

\begin{figure}[htbp]
        \centering
        \includegraphics[width = 0.8\linewidth]{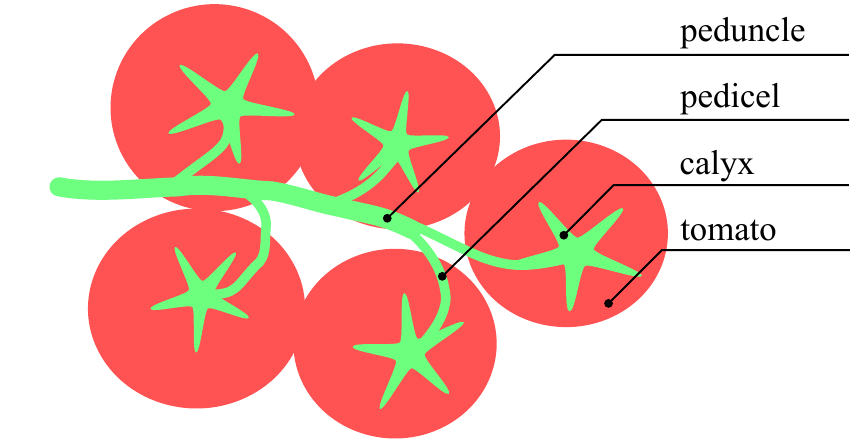}
        \caption{The vine tomato terminology: the stem consists of the peduncle, pedicels and calyxes combined.}
        % The ends of the blue curve are spread sufficiently far to prevent the truss from moving arbitrarily far from the yellow loop formed by the human hand.
        \label{fig:01:naming}
\end{figure}

In the sequel, we use the terminology according to Fig.~\ref{fig:01:naming}. The tomatoes are attached to the stem at the calyx, which is the flower-like structure. They are connected via fruit stalks, called pedicels, to the larger truss stalk called the peduncle. The pedicels are attached to the peduncle at locations which we refer to as junctions.

The rest of this paper is organized as follows: Section~\ref{sec:related_work} reviews the related work on grasping and peduncle detection of vine tomatoes. Section~\ref{sec:vision} presents the computer vision pipeline with the novel peduncle detection method that extracts the required geometric features of the tomato truss. Section~\ref{sec:grasping} describes the geometry-based grasping method. Vision and grasping lab experiments and their results are given in Section~\ref{sec:vision_experiments} and ~\ref{sec:experiments}, respectively. These results are further discussed in Section~\ref{sec:discussion}. Finally, Section~\ref{sec:conclusion} concludes this paper.

%%%%%%%%%%%%%%%%%%%%%%%%%
%%%%% RELATED WORK %%%%%%
%%%%%%%%%%%%%%%%%%%%%%%%%
\section{Related Work}
\label{sec:related_work}

While significant research has been done on the detection and harvesting of tomatoes \cite{bac2014harvesting}, only a few works focus on vine tomatoes. Vine tomato trusses cannot be grasped in the same way as single tomatoes, as grasping an entire truss by a single fruit would damage the tomato or detach it from the truss.

\subsection{Vine Tomato Grasping}

The PicknPack project developed a robotized production line to assess the product quality and pack fresh food products such as vine tomatoes \cite{pekkeriet2016picknpack}. A tomato truss is moved to a location in the harvest crate where the end effector has enough space to apply a pinch grasp on the fruit flesh. However, this way of grasping proved unsuitable, as it causes damage to the tomato skin.
Ji et al.~\cite{ji2014research} designed a harvesting robot for vine tomatoes in a greenhouse environment. After cutting, the truss is pinch-grasped with two rubber-plated fingers at the end of the peduncle. As a result, the grasped truss hangs almost vertically downwards, making accurate placement difficult. Kondo et al.~\cite{kondo2009machine} used a similar approach for grasping vine tomatoes.

As an alternative to the pinch-grasp methods, caging can be applied. Varava et al.~\cite{varava2016caging} determined sufficient caging conditions for rigid and partially deformable objects. Caging cannot guarantee a unique pose of the object and can only be used to manipulate objects without stringent precision requirements. To this end, Maeda et al.~\cite{maeda2012caging} equipped the end effector with soft pads that deform and apply a small reaction force to the object. Egawa et al.~\cite{egawa2015two} and Kim et al.~\cite{kim2019caging} analytically computed the required gripper distance to cage various rigid and deformable objects, such as two cubes connected via a rope, geometrically resembling a tomato truss.

\subsection{Stem Detection}
Kondo et al.~\cite{kondo2009machine} used the HSI (hue, saturation, intensity) color model to detect the connection of a truss to the tomato plant. Yoshida et al.~\cite{yoshida2019tomato} trained a support vector machine on RGB data to separate tomatoes from the background. Individual clusters and a cutting point on the peduncle were identified based on the geometry of the point cloud.
Benavides et al.~\cite{benavides2020automatic} combine edge detection with color-based segmentation to identify both the fruit and truss stem. The PicknPack project developed a vision system to detect vine tomatoes placed in a single non-overlapping layer in blue harvest crates \cite{pekkeriet2015d5}. Tomato trusses are segmented into stem and tomato flesh based on principle component analysis.
% Individual tomatoes are detected by fitting ellipses with a Hough transform.
% The fruit stem is detected based on its surface area and thickness.
The peduncle is identified using a \ac{ransac} regressor based on the assumption that the peduncle is the longest pixel area present in the stem segment. This assumption does not always hold, causing the peduncle classification to fail.

Recently, deep learning methods have been applied to tomato detection \cite{Liu20}.
However, no such methods have been developed for peduncle detection, perhaps because
peduncle labeling is much more involved and no suitable data set is readily available for training.
Hence, a traditional approach will be followed based on handcrafted features which gives more insight into the problem and allows for tuning by means of easy to interpret parameters.

%\todo[inline]{Taeke: Some sort of concluding remark?}

% Contrary to the research effort put in tomato harvesting and detection, not much literature exists on the grasping and identification of vine tomato. The three methods applied to vine tomato utilize a pinch based grasp, two at the end of the peduncle making the pose of the truss uncontrollable \cite{ji2014research, kondo2009machine}, and one at the skin of the tomato resulting in damage done to the fruit \cite{pekkeriet2016picknpack}. Where tomato detection has reached impressive results in complicated environment conditions, stem detection is mentioned only a few times. Only a single work identified the truss peduncle, but the approach fails for trusses where the peduncle does not make up the largest area of the stem \cite{pekkeriet2015d5}.

%%%%%%%%%%%%%%%%%%%%%%
%%%%%% VISION %%%%%%%%
%%%%%%%%%%%%%%%%%%%%%%
\section{Computer Vision Pipeline}
\label{sec:vision}
% \todo[inline]{Mention all parameters used?}

The computer vision pipeline extracts the geometric features of the tomato truss from RGB images. A setting is assumed, where a single truss is placed on a blue background with the stem upward. This particular lab setup is chosen to mimic the tomato packaging in a factory setting, where tomato trusses are first singulated and placed on a blue belt. These singulated trusses are later packaged manually. 
The pipeline steps are image processing, and tomato and peduncle detection, as further detailed below.
% % The hue angles are transformed to point on a circle with radius~\visparam{$r$}{radius of hue circle} since k-means clustering can not take into account the periodic nature of angular data. The a* color component is normalized such that is lies on the [-1, 1] range.
\subsection{Image Processing}
\label{image_processing}
\begin{figure}[htbp]
    \centering
    \includegraphics[width = 0.6\linewidth]{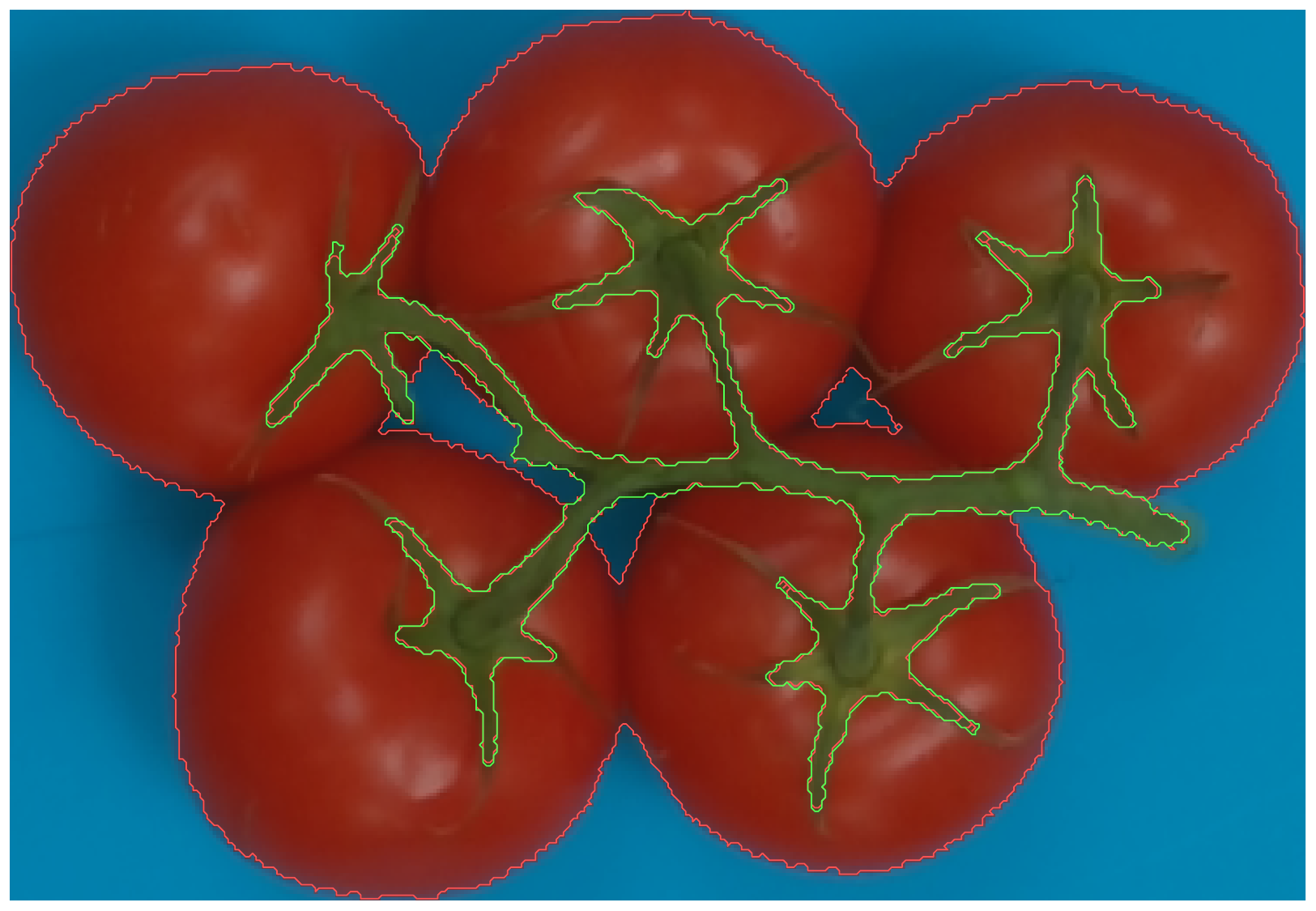}
    \caption{A cropped image overlaid with the class contours (for the sake of visualization only).}
    \label{fig:03:crop}
\end{figure}

Image processing consists of image segmentation, filtering, and cropping. First, the image is color-segmented such that each pixel is assigned to the background, stem (peduncle, pedicels, and calyxes), or tomato class. The a* color component from the L*a*b* color space and hue color component are computed for an image. The segmentation thresholds have been determined via k-means clustering.
%
%It is undesired to use the \acrfull{rgb} representation directly, since the separate color components are strongly correlated and share luminosity information together. Therefore, two alternative color components, hue and a*, are used for segmentation. The hue color component is an angular quantity, which starts at red, passes through green and blue, and wraps back to red. The a* color component, from the L*a*b* color space, distinguishes between green and red colors. K-means clustering is used to segment the image.
%
Subsequently, small pixel blobs and thin parts are removed using morphological opening and closing. Finally, the image is cropped using a rotated rectangle. The result is illustrated in Fig.~\ref{fig:03:crop}.

\subsection{Tomato Detection}
\label{tomato_detection}
\begin{figure}[htbp]
        \centering
        \includegraphics[width = 0.6\linewidth]{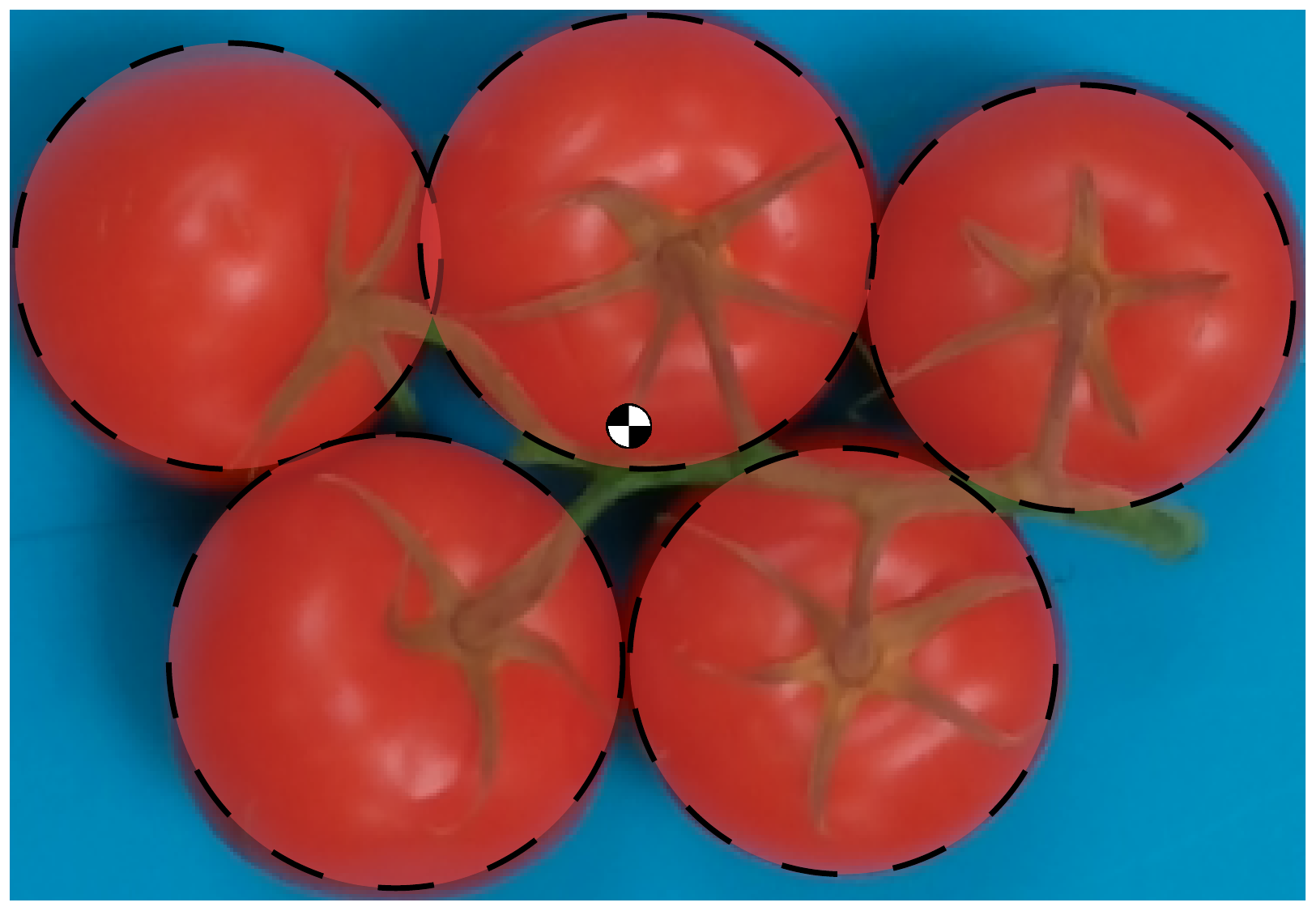}
        \caption{Result of the tomato detection step. The identified tomatoes are shown as the dashed circles, and the estimated truss center of mass is marked by the crossed circle.}
        \label{fig:03:detect_tomato}
\end{figure}

Tomatoes are modeled as circles, whose centers $c = (x, y)$ and radii $r$ need to be determined. First, the tomato segment edge is determined with the Canny algorithm. Than the circle Hough transform is used to fit circles to these edges. Several parameters are specified, such as the minimum and maximum tomato radius, minimum tomato center distance, resolution of the parameter space for the Hough transform, and sensitivities. False positives are removed by discarding all circles which do not overlap the tomato segment by at least 50\%. For each tomato, its center of mass is estimated by assuming that the stem mass is negligible compared to the tomato mass, and that the tomato mass scales with the volume of a sphere with an identical radius~\cite{nyalala2019tomato}. Consider a truss with $N$ tomatoes. If the $i^\text{th}$ tomato has a center $c_i$ and a radius $r_i$, then the center of mass of the truss is computed as:
\begin{equation}
    c_{\text{com}} = \frac{ \sum\limits_{i = 1}^{N} r_i^3  c_{i}}{\sum\limits_{i = 1}^{N} r_i^3}.
\end{equation}
%  are the center and the radius of the $i^{th}$ tomato, respectively
The computed center of mass of the entire truss is later used to determine a suitable grasping point. The result is illustrated in Fig.~\ref{fig:03:detect_tomato}.

\subsection{Peduncle Detection}
\label{penduncle_detection}
\begin{figure}[htbp]
        \centering
        \begin{minipage}{0.6\linewidth}
            \centering
            \vskip 0pt
            \subfloat[Skeletonization of the stem segment.
            \label{fig:03:peduncle:a}]{
                \includegraphics[width=\linewidth]{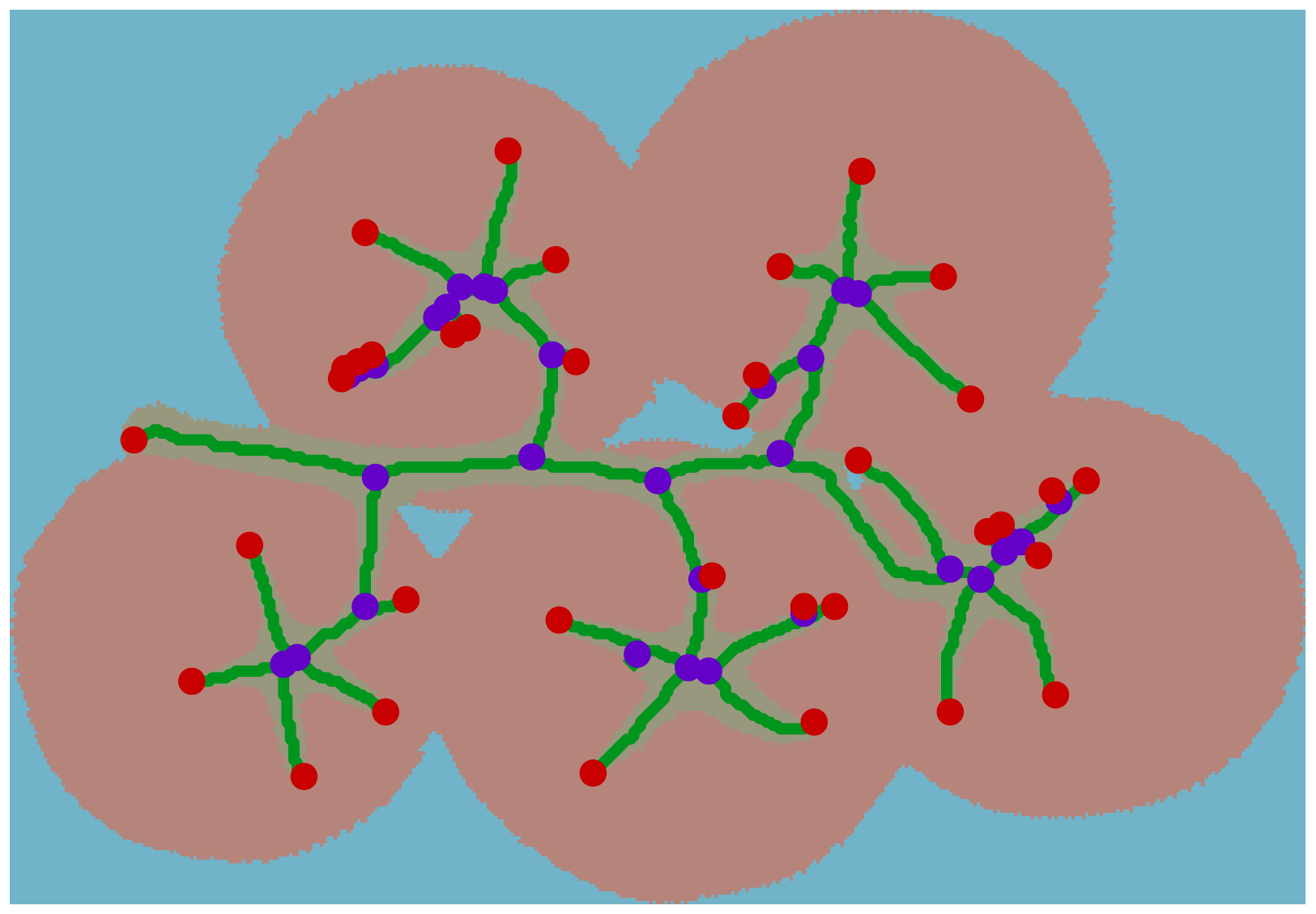}
        }
    \end{minipage}
        \begin{minipage}{0.6\linewidth}
            \centering
            \vskip 0pt
        \subfloat[Distance based thresholding of junction-tail type branches. \label{fig:03:peduncle:b}]{
                \includegraphics[width=\linewidth]{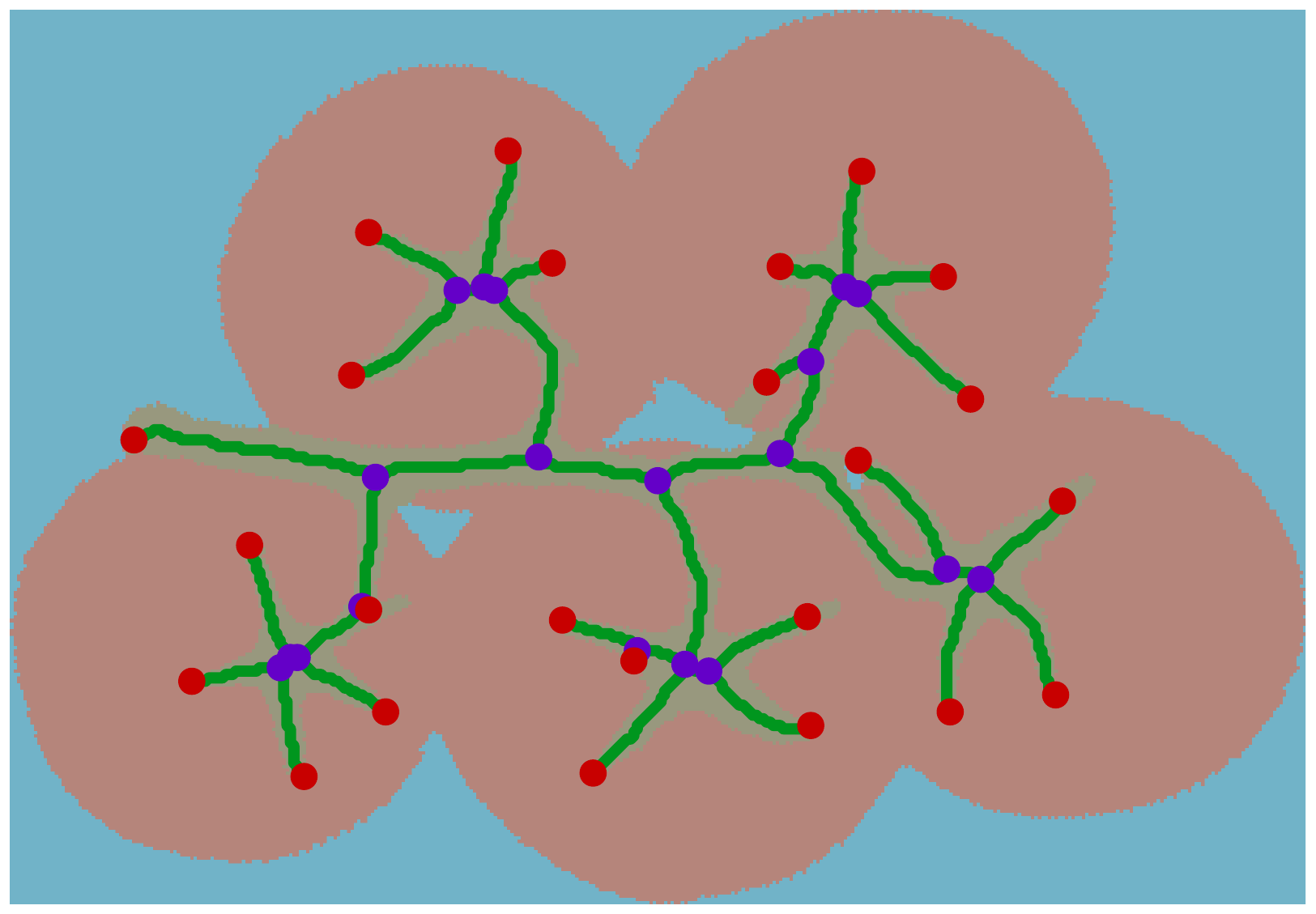}
        }
    \end{minipage}
        \begin{minipage}{0.6\linewidth}
            \centering
            \vskip 0pt
        \subfloat[Identified peduncle. \label{fig:03:peduncle:c}]{
                \includegraphics[width=\linewidth]{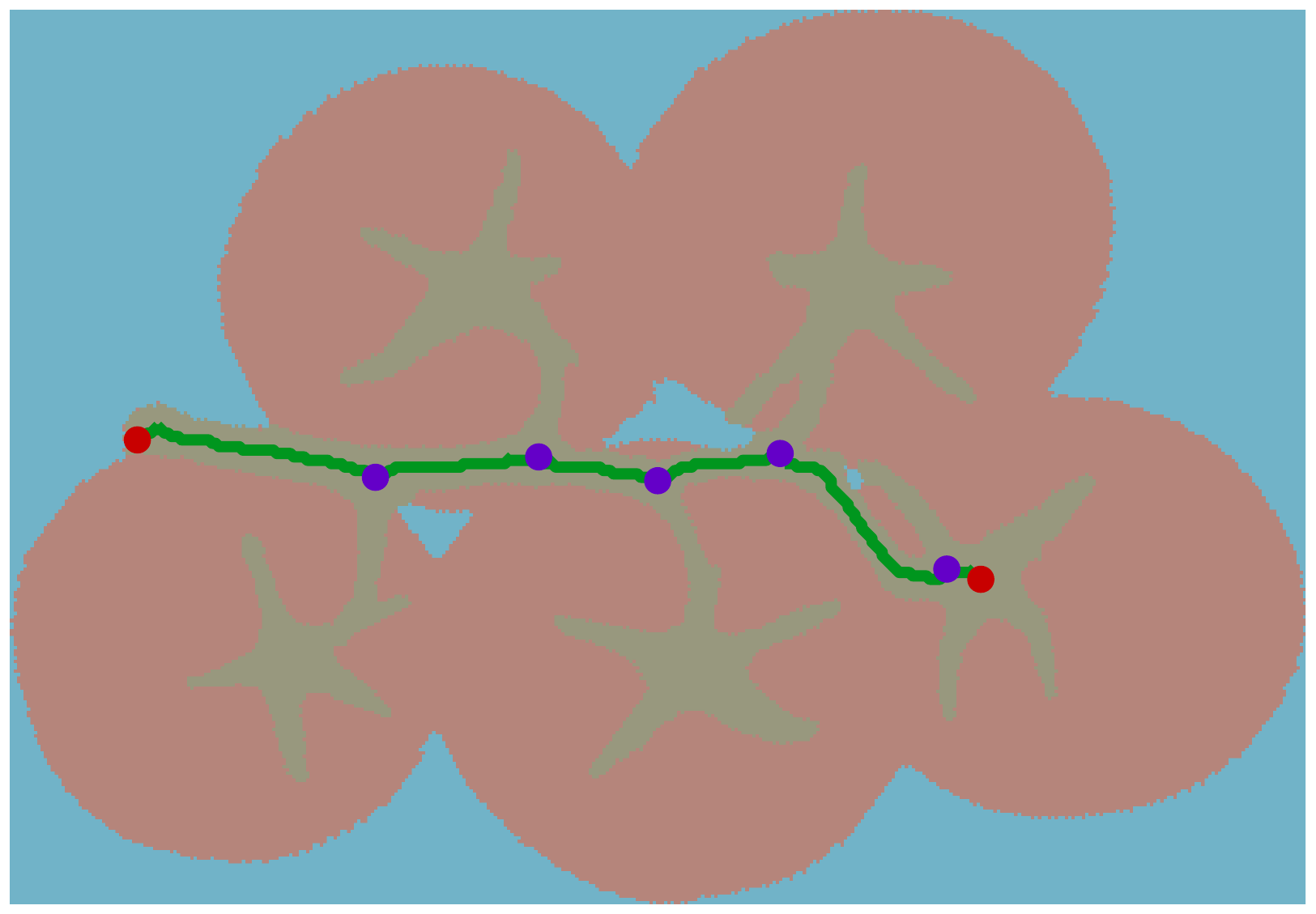}
        }
    \end{minipage}
    \caption{The peduncle detection process. Edges are marked by the dark green lines, junctions by the purple dots, and tails by the red dots.}
        \label{fig:03:peduncle}
\end{figure}

To distinguish the peduncle from the other stem components, a graph-based method is employed. First, a single-pixel-wide representation of the stem segment is computed and then it is converted to a graph. The different stem branches are represented as edges interconnected via vertices. Each vertex $v$ is represented by its 2D location $(x, y)$ on the image. Each edge is associated with a real-valued weight that represents the length measured along the edge. The degree of a vertex is the number of edges incident with this vertex. A vertex is either called a tail when its degree is one, or a junction when its degree is three or more. A truss overlaid with its graph is shown in Fig.~\ref{fig:03:peduncle:a}.
%A pixel connected to a single other pixel is called a tail vertex. If at least two other pixels are connected it is called a junction vertex. A skeleton segment interconnecting two vertices is called an edge.\RB{These are some definitions of what is what, but it is not explained how the graph is constructed.}
%
%
Spurious short edges can result from the stem skeletonization. These are removed by thresholding the junction-tail type edges at a minimum length of 10\,mm, as shown in Fig.~\ref{fig:03:peduncle:b}.

To find the peduncle, we determine the longest path on the graph with a limited curvature. 
A path is a sequence of alternating vertices and edges, all distinct, and its length is given by the sum of its edges' lengths. The orientation $\theta_e$ of an edge is computed using the edge's source vertex $v_{s} =  (x_{s},\, y_{s})$, and the destination vertex $v_{d} = (x_d,\, y_d)$ as:

\begin{equation}
    \theta_{e} =  \arctan(\frac{y_d - y_s}{x_d - x_s}).
\end{equation}
Similarly, the orientation of an entire path $\theta_{p}$ is computed using the origin and terminus vertex of the path.
%The orientation of an edge is computed by taking the arctangent of the difference between the origin and terminal vertex of the edge. The angle of an entire path is computed using the origin and terminal vertex of the path.
%The curvature between an edge and a path is approximated by the difference between their orientation. % The peduncle is identified from the graph by searching for the path with the largest weight, where a path is split if the curvature exceeds a threshold.
The curvature between an edge and the path is the absolute difference between their orientations: $c = |\theta_{p} - \theta_{e}|$.

To find the peduncle, the graph is traversed starting from any arbitrary tail vertex towards any other tail vertex. Each newly generated path is compared to the longest path seen so far. If the length of the new path exceeds the length of the longest path, then the longest path is updated. 
When, at a certain vertex, the orientation between the path traversed so far and the succeeding edge exceeds the threshold of $45^\circ$, then the path is split at this vertex. If another succeeding edge that lies within the curvature threshold is found at this vertex, then the original path continues. Otherwise a new path is started from this vertex and the original path is stored if it is longer than the stored longest path. %This means that This particular path could not find a tail-to-tail connection with a limited curvature. 
%Meaning that the current path is finished, and a new path is started with initially zero length. 
Once this search is performed starting from all tail vertices, the longest path found is the peduncle. The resulting peduncle with junctions and tails is shown in Fig.~\ref{fig:03:peduncle:c}.
Note that a peduncle does not have to be a tail-to-tail path.

%%%%%%%%%%%%%%%%%%%%%%%%%
%%%%%%% GRASPING %%%%%%%%
%%%%%%%%%%%%%%%%%%%%%%%%%
\section{Geometry-Based Grasping}

\label{sec:grasping}

This section describes a geometry-based grasping method for vine tomatoes. For a successful grasp, we assume that: \begin{inparaenum}[(i)]\item the truss is placed on a horizontal surface with the peduncle facing upward, \item the target truss is separated from other trusses, and \item sufficient space is available above the truss for the manipulator to approach it from above.\end{inparaenum}
%These assumptions are reasonable if the trusses are taken out of the transportation crates.

\subsection{Geometric Model of the Vine Tomato}
\label{sec:geometric_model}

\begin{figure}[htbp]
        \centering
        \includegraphics[width = 0.8\linewidth]{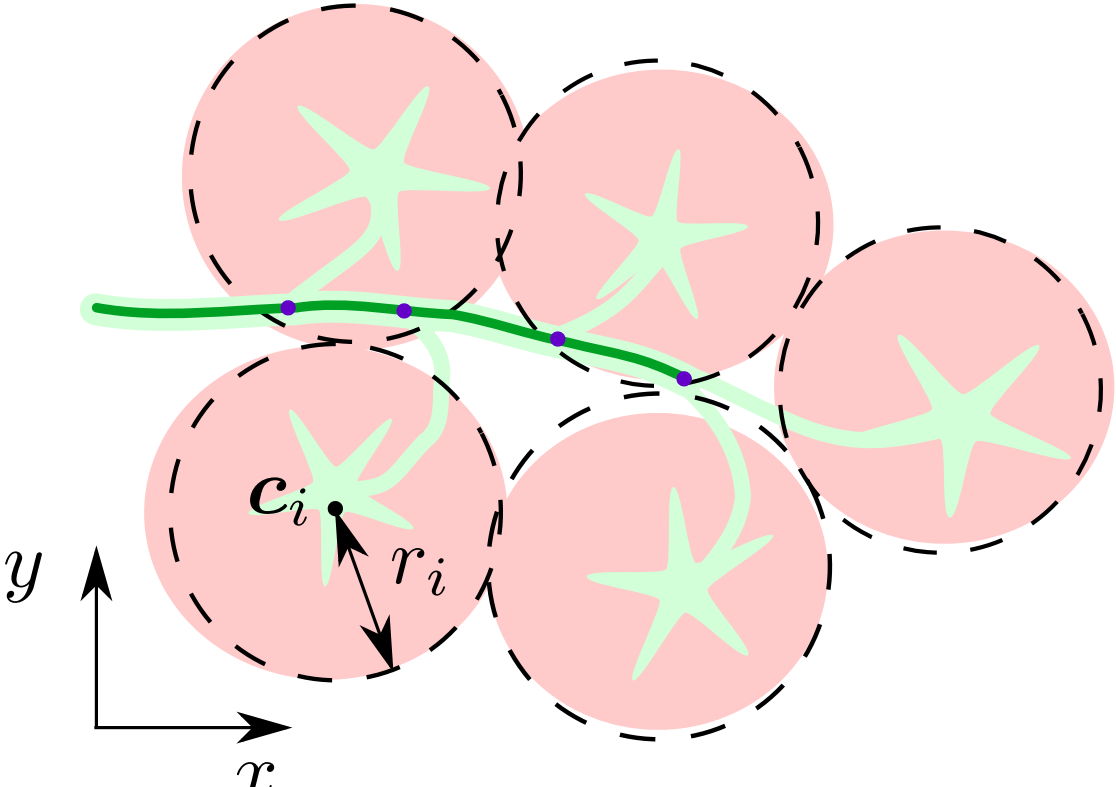}
        \caption{A vine tomato overlapped with its geometric model. The tomatoes are represented by the dashed circles, the peduncle by the dark green curve, and the junctions by the purple dots.}
        \label{fig:02:truss_model}
\end{figure}

We use a two-dimensional geometric model of the truss and of the end effector to determine a suitable grasp location and gripper orientation. Tomato $i$ is modeled as a circle with radius $r_i$ and center $c_i \in \R^2$, see Fig.~\ref{fig:02:truss_model}.

\begin{figure}[htbp]
        \centering
        \begin{minipage}{0.49\linewidth}
            \centering
            \subfloat[Side view \label{fig:02:ee_model:1}]{
                \includegraphics[height=4.5cm]{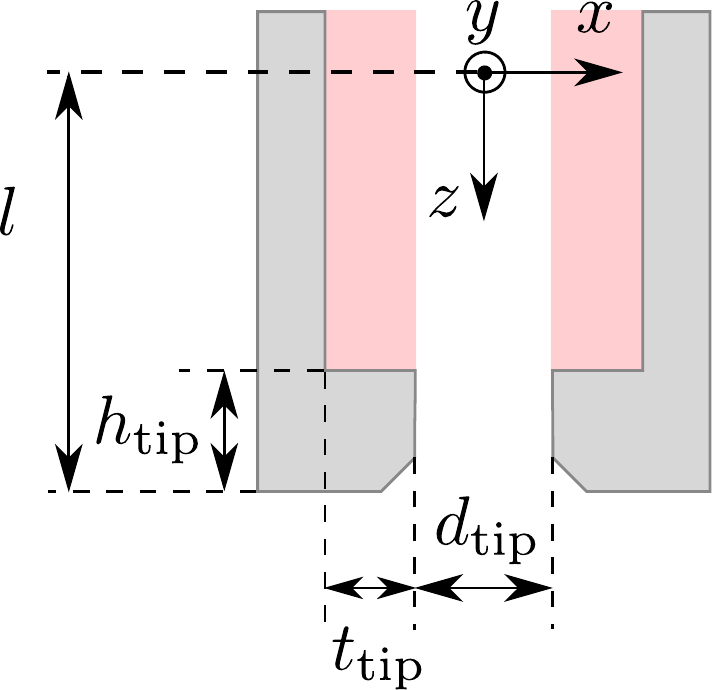}
        }
    \end{minipage}
        \hfill
        \begin{minipage}{0.49\linewidth}
            \centering
        \subfloat[Front view \label{fig:02:ee_model:2}]{
            \includegraphics[height=4.5cm]{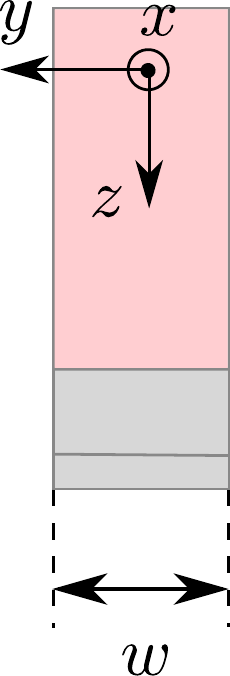}
        }
    \end{minipage}
        \caption{Geometric model of the end effector, shown from two different views. The red area marks the deformable material.}
        \label{fig:02:ee_model}
\end{figure}

%\todo[inline]{Taeke: Maybe only mention finger width $w$ and finger tip distance $d_\text{tip}$ since there are only relevant parameters.}
The end effector is a parallel gripper of length $l$ and width $w$, see Fig.~\ref{fig:02:ee_model}. The fingertips of the L-shaped fingers are of height $h_\text{tip}$ and thickness $t_\text{tip}$. The distance between the finger tips $d_\text{tip}$ can be controlled. The inner surface of the fingers is covered with soft material to allow for more precise control of the post-grasp truss pose.
%This is required to meet the caging-based grasping conditions as discussed next.
%

\subsection{Grasp Constraints}
\label{sec:geometric_constraints}

The geometric models are used to derive constraints that represent the caging conditions. % derive constraints for the finger tip distance and grasp pose to meet the caging-based grasping conditions.
% A caged object can move freely in the closed region unless the region is a single point, which is called form closure.
Maeda et al.~\cite{maeda2012caging} introduced caging-based grasping with deformable parts on the gripper to apply a small reaction force to the object. The following two constraints need to be satisfied for a caging-based grasp:
\begin{enumerate}
        \item \textbf{Rigid-part caging condition}: the target object is in a complete imprisonment of the rigid parts of the end effector.
        \item \textbf{Soft-part caging condition}: assuming that the soft parts of the end effector are rigid, the closed region for caging in the configuration space of the object becomes empty.
\end{enumerate}
The former condition implies that the object cannot escape from the end effector, while the latter condition implies that the soft parts of the gripper deform, and therefore apply a reaction force to the target object. Both conditions can be tested geometrically.

The rigid-part caging condition can be verified by noting that the peduncle exhibits various double fork features, due to the pedicels attached. It is assumed that these features are preserved under deformations of the stem. Therefore, the rigid-part caging condition is met when the end effector encloses a section of the peduncle with junctions on each side.
The soft-part deformation condition is satisfied when the diameter of the peduncle is larger than the fingertip distance; this requirement is automatically satisfied when the end effector is fully closed.

%The geometry of the tomatoes is not used, thus a cage may be applied to a truss even when tomatoes are missing.

% and there needs to be sufficient space for the end effector fingers. Generally many poses exist which satisfy these requirements, the pose closest to the center of mass is chosen is selected to prevent tilting.

To determine a unique target grasp location on the peduncle, we introduce two additional conditions:
\begin{itemize}
        \item \textbf{Space condition}: the target grasp location lies at least at a distance $L$ from the junctions.
        \item \textbf{Balance condition}: The target grasp location lies as close as possible to the truss' center of mass.
\end{itemize}
The space condition is introduced such that there is sufficient space for the end effector on the target grasp location.
The distance $L$ is taken larger than the end effector's width, to allow for small errors in manipulator control.
The balance condition is introduced to prevent undesired tiling of the truss. The selected grasping point is the candidate location which lies closest to the center of mass.

\subsection{Grasp Pose}
\label{sec:grasping_strategy}

\begin{figure}[htbp]
        \centering
        \includegraphics[width = 0.8\linewidth]{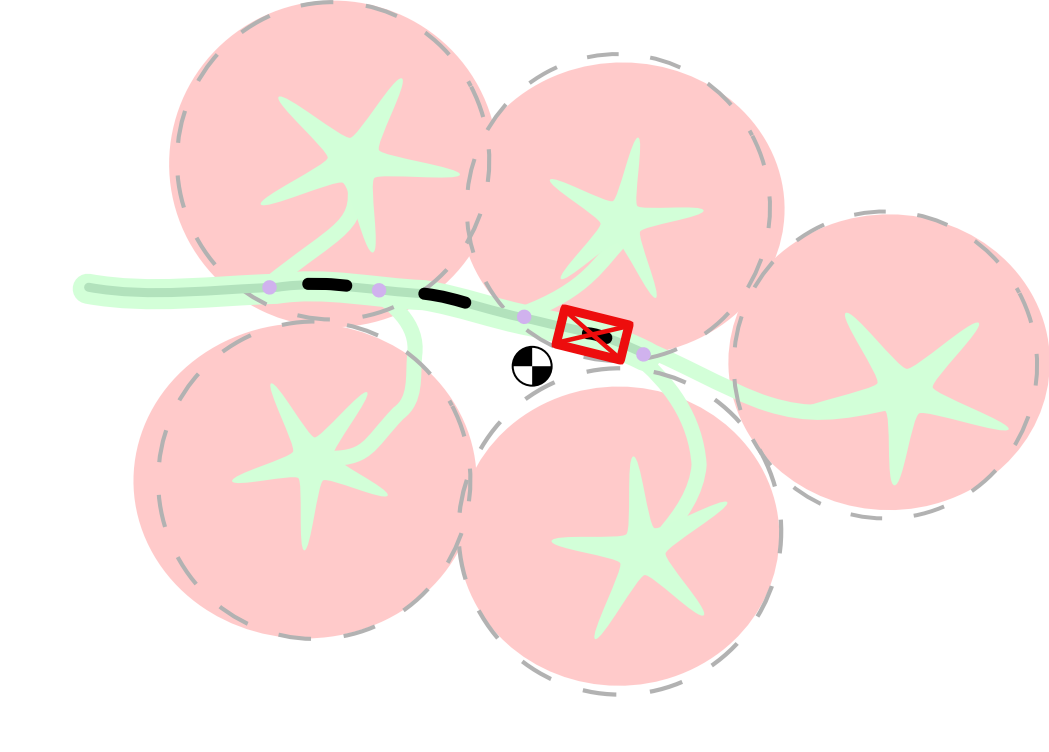}
        \caption{The geometric model overlapped with the determined grasping location. The center of mass is shown as a crossed circle, the possible grasping locations are the black lines, and the target grasping location is shown by the red box.}
        \label{fig:02:grasp}
\end{figure}

The end effector is aligned with the peduncle.
The target grasp pose for the previously discussed truss is shown in Fig.~\ref{fig:02:grasp}. This two dimensional grasp pose is transformed to 3D by adding the grasp height, which is derived from the distance of the peduncle above the surface and the end effector dimensions.

To grasp a truss, the robot's controller executes the following procedure. First, the manipulator moves to a pre-grasp location which is above the target grasp location. The robot then moves the end effector downward to the grasp pose and closes the gripper. Once the caging-based grasp is accomplished, the robot can lift the truss and place it at a desired location.

\section{Computer Vision Experiments}
\label{sec:vision_experiments}
The computer vision pipeline was tested on a dataset consisting of 84 labeled images, containing 308 tomatoes and 266 junctions. We counted the number of true positive detections (TP), false positive detections (FP), and false negative detections (FN). The true positive rate (TPR) and false discovery rate (FDR) are computed, which describe the ability of the pipeline to find truss features and its proneness to classify negative examples as positive ones:
\begin{equation}
        \text{TPR} = \frac{\text{TP}}{\text{TP} + \text{FN}}, \hspace{10mm}
        \text{FDR} = \frac{\text{FP}}{\text{FP} + \text{TP}}
\end{equation}
Furthermore, the errors of the estimated features are computed and reported as the mean absolute error plus and minus the standard deviation.

\subsection{Results}
For the tomato detection we obtained a true positive rate of 99.7\% (307/308). No false positives were reported. The tomato centers were predicted with an error of 3.58 $\pm$ 2.63mm (n=307) and the radii were estimated with an error of 2.43 $\pm$ 2.21mm (n=307). As a result, the center of mass is estimated with an error of 5.02 $\pm$ 3.26\,mm~(n=307). For the junction detection, we obtained an overall true positive rate of 86\% (229/266) and a false discovery rate of 34\% (117/346). The junction locations were predicted with an error of 2.09 $\pm$ 1.78mm (n = 229). 

\begin{figure}[htbp]
        \centering
        \begin{minipage}{\linewidth}
            \centering
            \vskip 0pt
            \subfloat[True positives are marked by the dashed circles, and the determined center of mass by the crossed circle. The values state the error made in determined location (loc) and radius (r).
            \label{fig:04:result:tomato}]{
                \includegraphics[width=\linewidth]{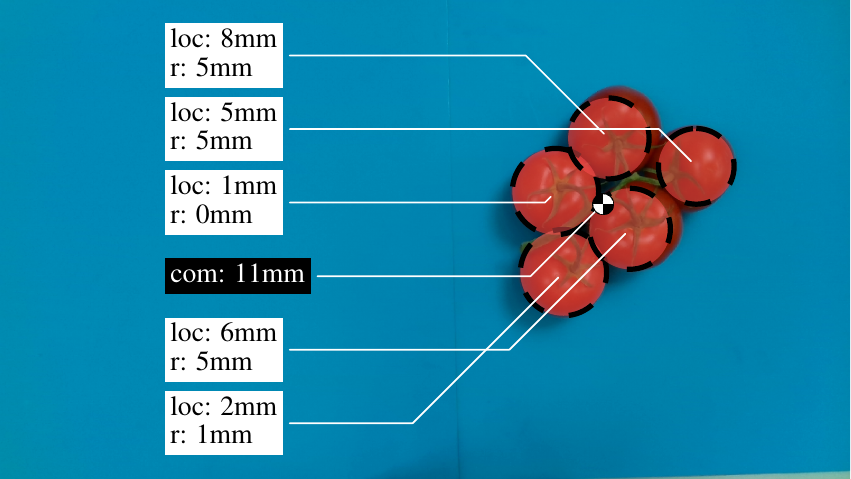}
        }
    \end{minipage}
        \begin{minipage}{\linewidth}
            \centering
            \vskip 0pt
        \subfloat[True positives are marked by the purple dots, the corresponding value states the error made in determined location (loc). \label{fig:04:result:peduncle}]{
                \includegraphics[width=\linewidth]{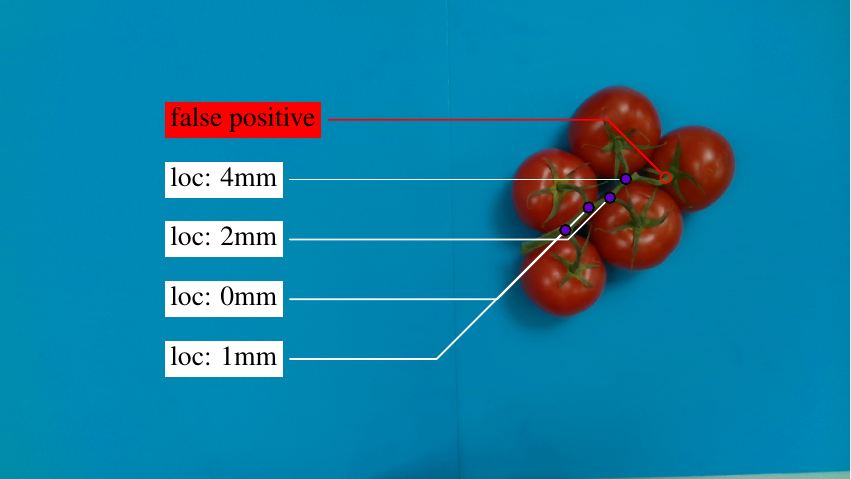}
        }
    \end{minipage}
    \caption{Illustration of typical results obtained for the computer vision pipeline.}
    \label{fig:04:result}
\end{figure}

\begin{figure}
    \centering
    \includegraphics[width = \linewidth]{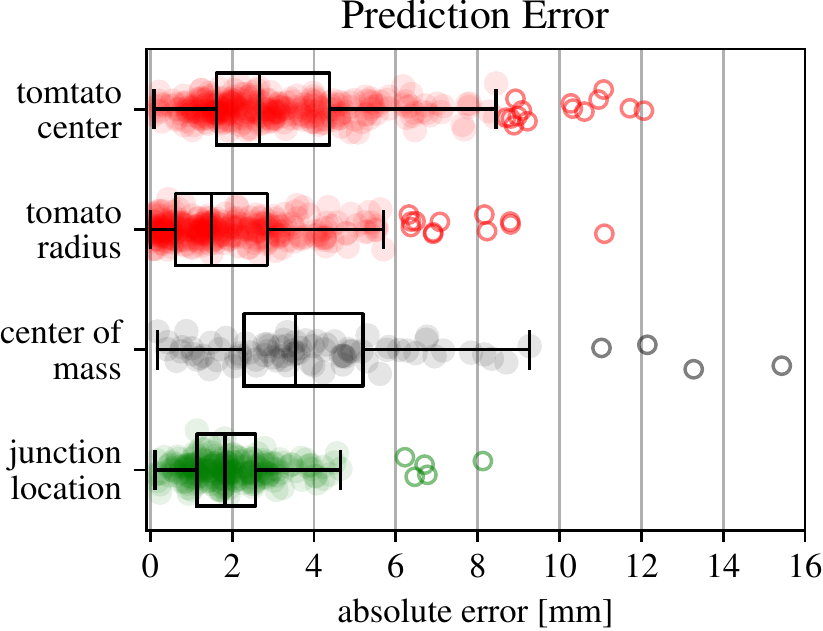}
    \caption{Boxplot displaying the absolute prediction error of the computer vision pipeline for several truss features.}
    \label{fig:box_plot}
\end{figure}

Typical tomato and junction detection results are shown in Fig.~\ref{fig:04:result:tomato} and Fig.~\ref{fig:04:result:peduncle}, respectively. A box plot of the prediction errors for the different truss features is shown in Fig.~\ref{fig:box_plot}. Based on the obtained features, the algorithm was able to find a target grasp pose on 96\% (81/84) of the images. The computer vision pipeline took 1.88 $\pm$ 0.83 seconds to execute per image. Peduncle detection, cropping and segmentation take up a significant amount of time.

\subsection{Discussion}
%\todo[inline]{Taeke: Do we want to discuss the result of tomato detection. It is not so interesting since it has been done many times before on much more complicated data sets.}
% The pipeline is able to identify almost all tomatoes without reporting any false positives. The center of mass estimate is sensitive to the prediction errors of the tomato features.
%
Two factors affect this prediction error. First of all, the tomato features are harder to estimate when a tomato is closely surrounded by other tomatoes or when it is at the edge of the image. For these tomatoes, the circle can only be fitted to a small part of its edge. Secondly, the predictions are affected by the ellipsoidal shape of the tomatoes; sometimes the circle is fitted to a part of the edge which has a smaller or larger radius than the tomato.

The proposed graph-based peduncle search method is able to correctly identify the peduncle in cases where the previously proposed \ac{ransac} regressor fails \cite{pekkeriet2015d5}. The newly developed method works best when the camera is placed directly above the truss, the performance degrades when the object is placed off-center.
Many junctions are correctly detected, but also false positives are reported.
The junction detection is sensitive to overlapping parts of the stem and to small segmentation errors. Furthermore, the graph-based peduncle search sometimes classifies a pedicel and parts of a calyx as peduncle, resulting in false positives.

%\todo[inline]{Taeke: Need to add example that peduncle detection works in cases where RANSAC fails.}
%\todo[inline]{Taeke: Do I need a conclusion here? Maybe strange to do in the middle of a paper?}

%%%%%%%%%%%%%%%%%%%%%%%%%%
%%%%%% EXPERIMENTS %%%%%%%
%%%%%%%%%%%%%%%%%%%%%%%%%%
\section{Robotic Grasping Experiments}
\label{sec:experiments}
We have conducted real-world experiments to validate the proposed geometry-based grasping method for vine tomatoes. These experiments were performed in a lab environment.

\subsection{Experimental Setup}
\begin{figure}[htbp]
        \centering
        \begin{minipage}{0.64\linewidth}
            \centering
            \vskip 0pt
            \subfloat[Overview
            \label{fig:05:setup:a}]{
                \includegraphics[height = 3.2cm]{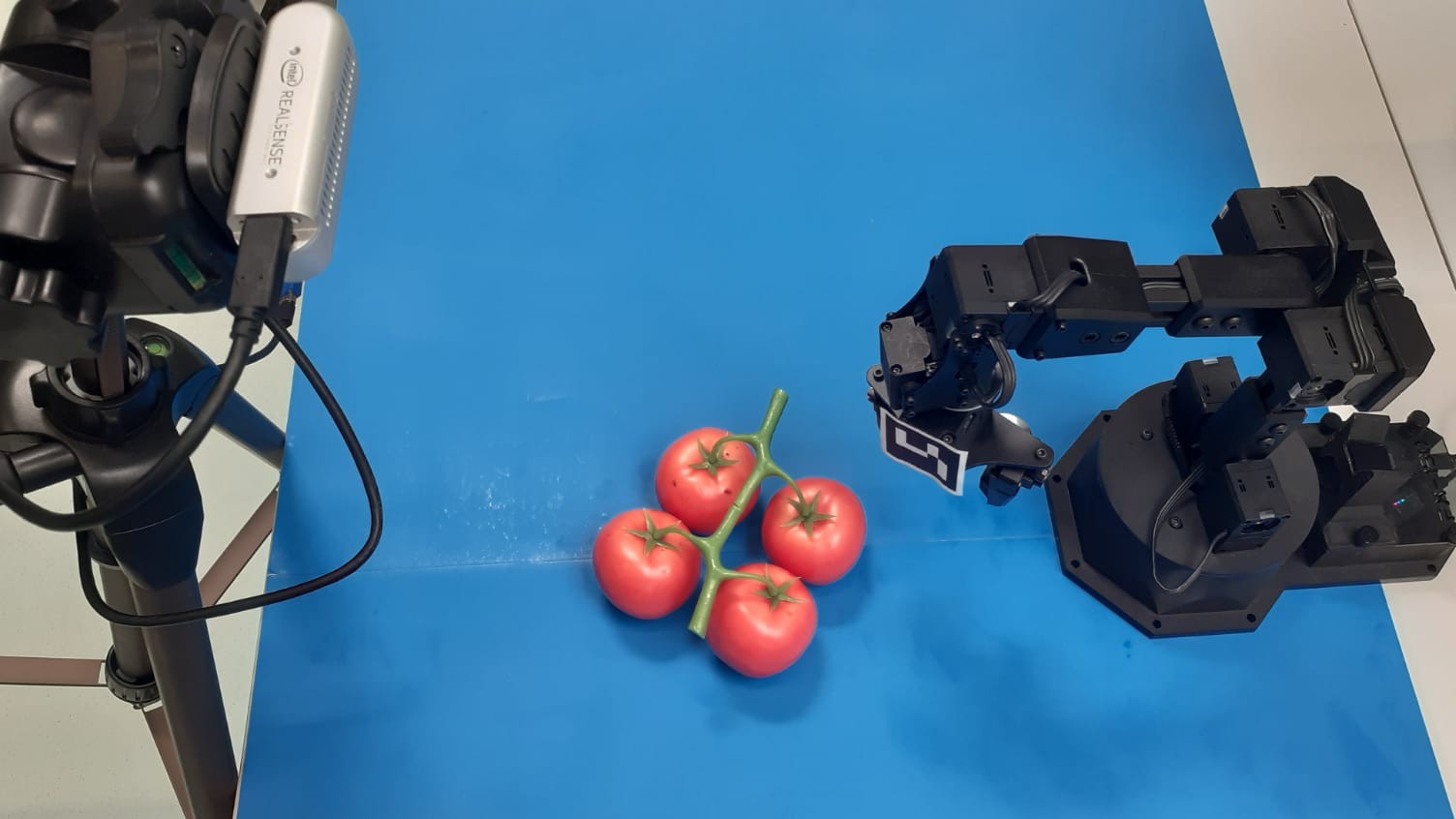}
        }
    \end{minipage}
    \hfill
        \begin{minipage}{0.34\linewidth}
            \centering
            \vskip 0pt
        \subfloat[End effector \label{fig:05:setup:b}]{
                \includegraphics[height = 3.2cm]{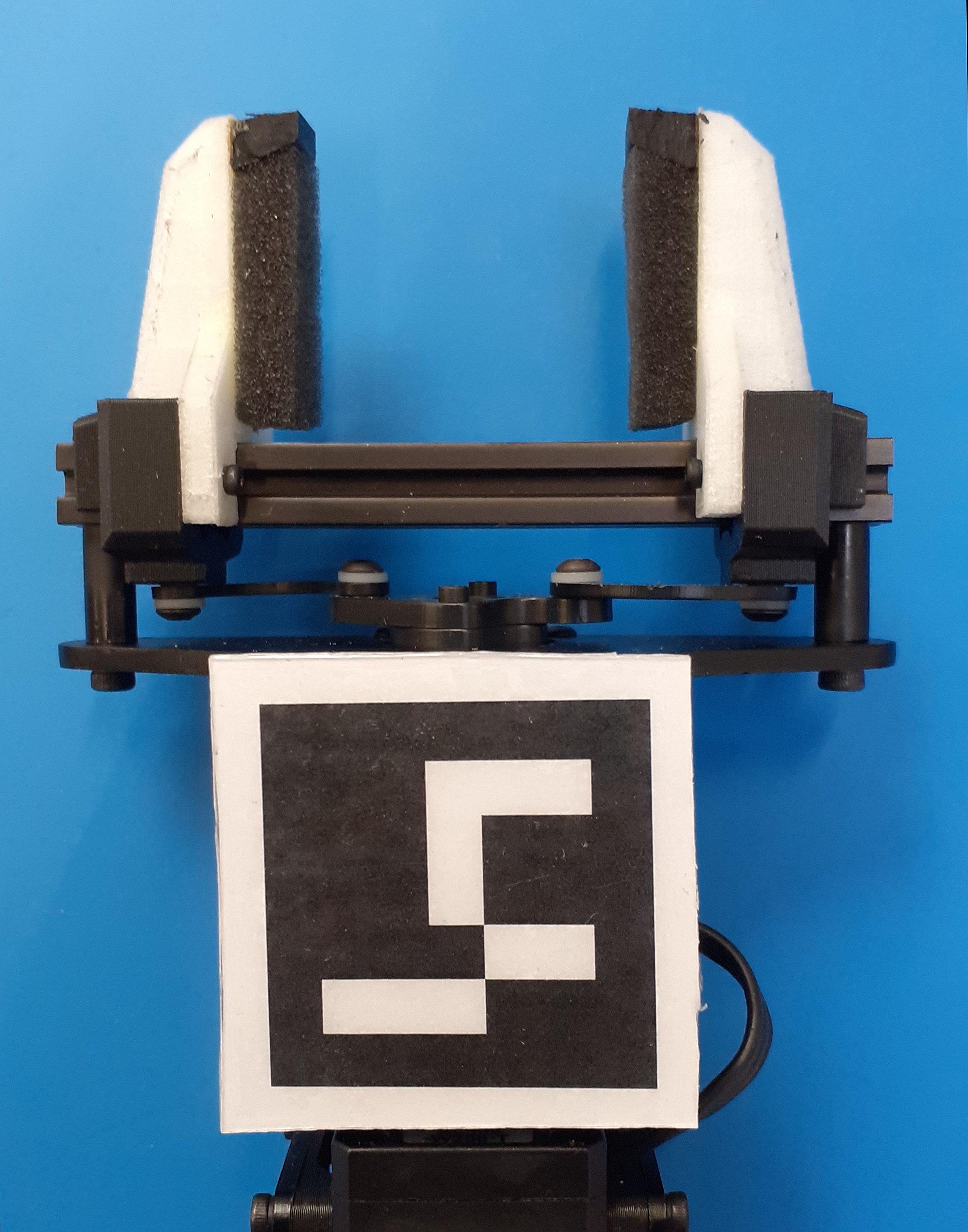}
        }
    \end{minipage}
        \caption{Overview of the experimental setup (a), and the end effector (b).}
        \label{fig:05:setup}
\end{figure}

The physical setup consists of a robotic manipulator, an RGB-D vision system and several tomato trusses as shown in Fig.~\ref{fig:05:setup:a}. The Interbotix PincherX 150 Robot Arm has five degrees of freedom\footnote{\url{https://www.trossenrobotics.com/pincherx-150-robot-arm.aspx}}. The gripper fingers are covered with deformable foam as required for the soft-part caging condition. Furthermore, rubber knobs were attached to these fingers such that they can enclose the peduncle. Fig.~\ref{fig:05:setup:b} shows the end effector in its open configuration.

We used the Intel RealSense Depth Camera D435\footnote{\url{https://www.intelrealsense.com/depth-camera-d435/}} which combines active \ac{ir} stereo technology to measure depth with an RGB sensor for color information. Since a sense-plan-act framework is used, the camera is placed in an eye-on-base configuration. The camera in unable to obtain accurate depth information about the thin peduncle. Hence, the peduncle height is taken as a constant measured before conducting an experiment. Our code is available on GitHub.\footnote{\url{https://github.com/padmaja-kulkarni/taeke_msc}}

%\todo[inline]{Taeke: What truss stems do we want to use for the realistic model(s), maybe redo some experiments?}
Due to the 50g payload limitation of the manipulator, we had to use plastic tomato models. Six trusses of two different categories were created:
\begin{enumerate}
        \item \textbf{Simple truss}: plastic tomatoes are attached to a plastic peduncle. The peduncle has a relatively large diameter and a simple straight shape. % Three models were created as shown in Figure~\ref{fig:05:target:a} to \ref{fig:05:target:c}.
        \item \textbf{Realistic truss}: plastic tomatoes are attached to a real tomato peduncle. Truss stems with sufficient space around the peduncle were selected to accommodate the gripper fingers. % Trusses of various shapes were created, as shown in Figure~\ref{fig:05:target:d} to \ref{fig:05:target:f}.
\end{enumerate}

\subsection{Task}
A pick and place routine is executed to determine the strengths and weaknesses of the method. The task consists of five successive steps: (i) the truss is detected and the relevant features are extracted, (ii) the end effector moves toward the peduncle and grasps it, (iii) the truss is lifted off the supporting surface, (iv) the truss is moved to an arbitrary target pose, and (v) the truss is placed back on the supporting surface. This task is successful if all these parts are completed. The following modes of failure were observed:
\begin{itemize}
        \item \textbf{Vision}: the vision system fails to identify the key features required for the geometry-based grasping, such as the peduncle, or the junctions.
        \item \textbf{Planning}:  the grasp planning algorithm fails to determine a valid grasp action. For example, it may require a pose that cannot be reached by the end effector due to space constraints. This error indicates weaknesses in the geometry-based grasping method.
        \item \textbf{Control}: the manipulator fails to execute the planned task. A typical example is the end effector not reaching the target pose. These errors show weaknesses in the hardware, calibration, and control implementation.
        \item \textbf{Overload}: the hardware fails to execute the task at hand due to overloading or overheating of the servo motors.
\end{itemize}
In case the task is not successfully executed, the mode of failure will be recorded for later discussion. Since the goal of this experiment is not to test the load capacities of the robot, overload failures will be ignored. When a grasp fails the truss is moved to a new random pose by the human operator such that the next experiment can continue.

\subsection{Results}
\label{sec:05:results}
\begin{figure}[htbp]
    \vspace{4mm}
    \centering
    \begin{minipage}{\linewidth}
        \centering
        \vskip 0pt
        \subfloat[A successful grasp where the end effector fingers wrap around the peduncle.
        \label{fig:06:result:a}]{
                \includegraphics[height=3.8cm]{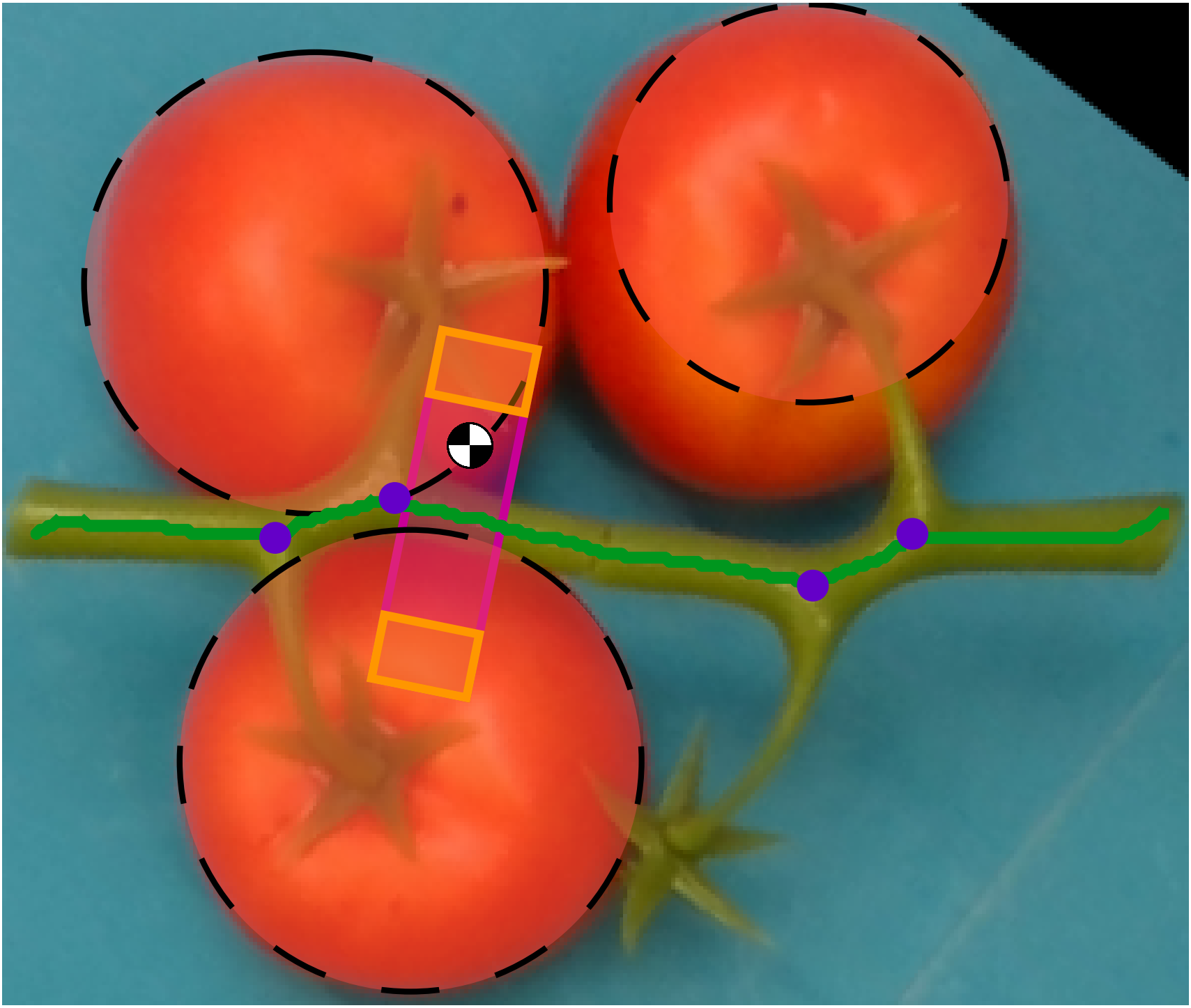}
                \hfill
                \includegraphics[height=3.8cm]{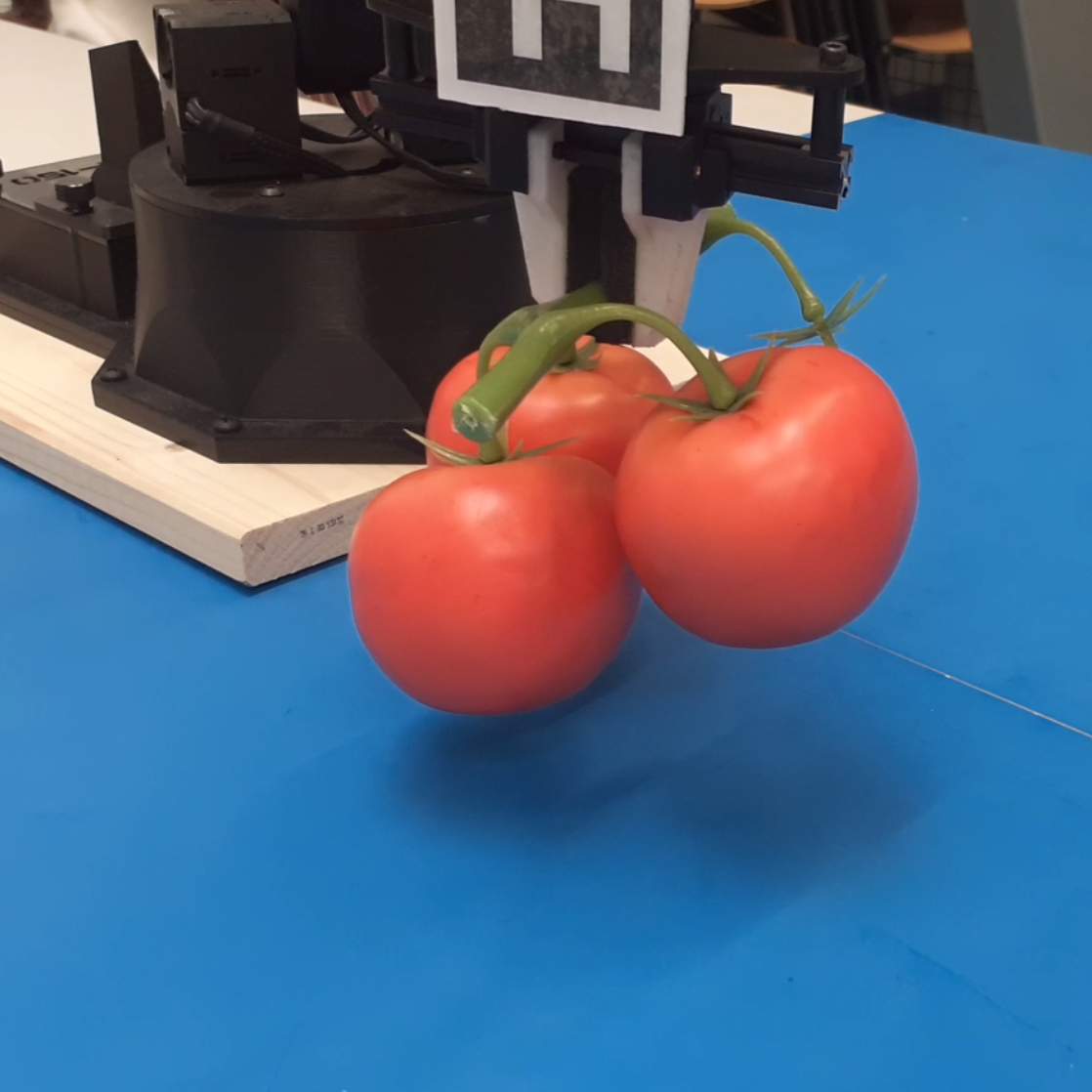}
        }
    \end{minipage}
    \hfill
        \begin{minipage}{\linewidth}
            \centering
            \vskip 0pt
        \subfloat[A failed grasp where the planned closing actions causes the end effector fingers to get stuck on the pedicel. \label{fig:06:result:b}]{
                \includegraphics[height=3.8cm]{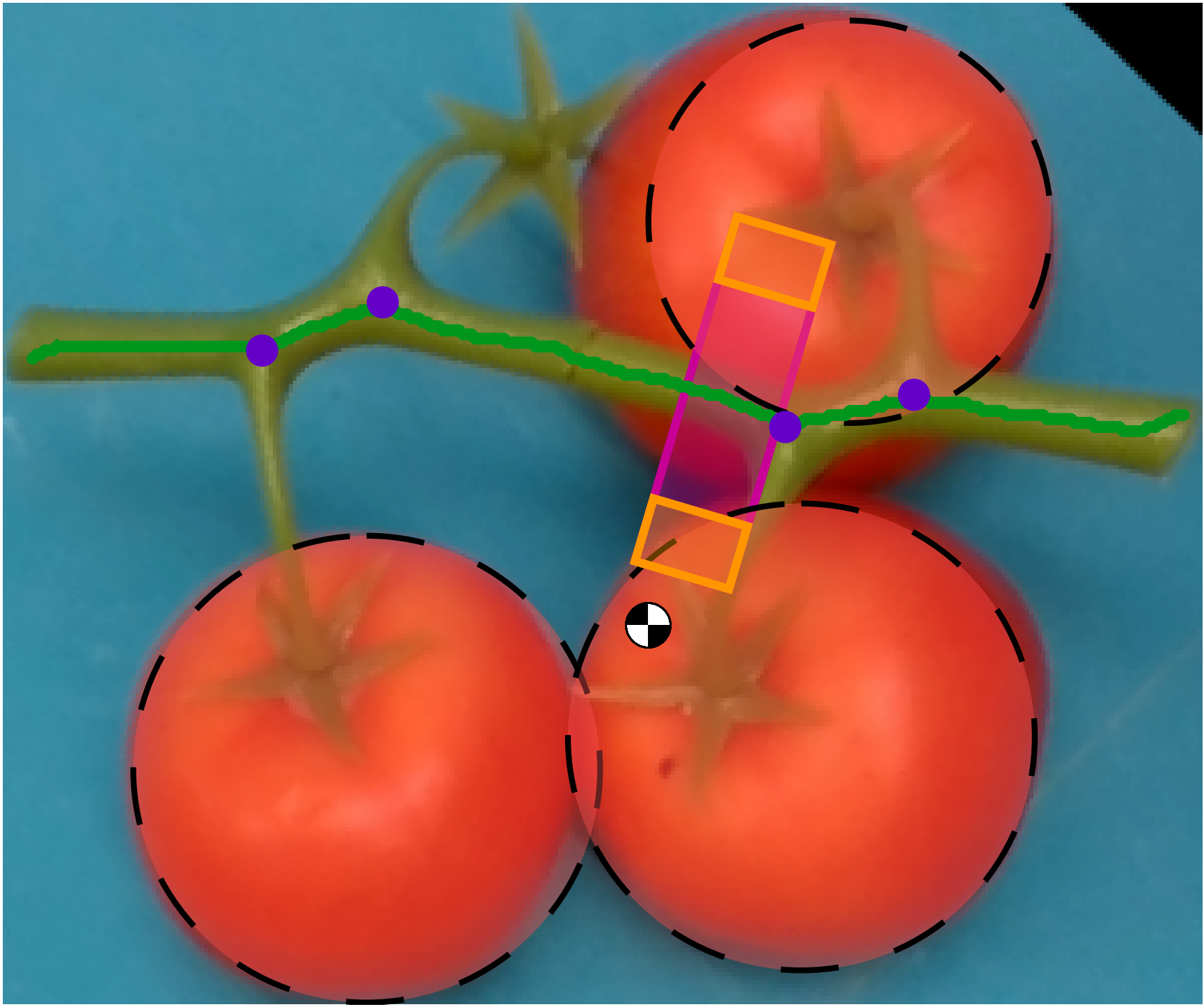}
                \hfill
                    \includegraphics[height=3.8cm]{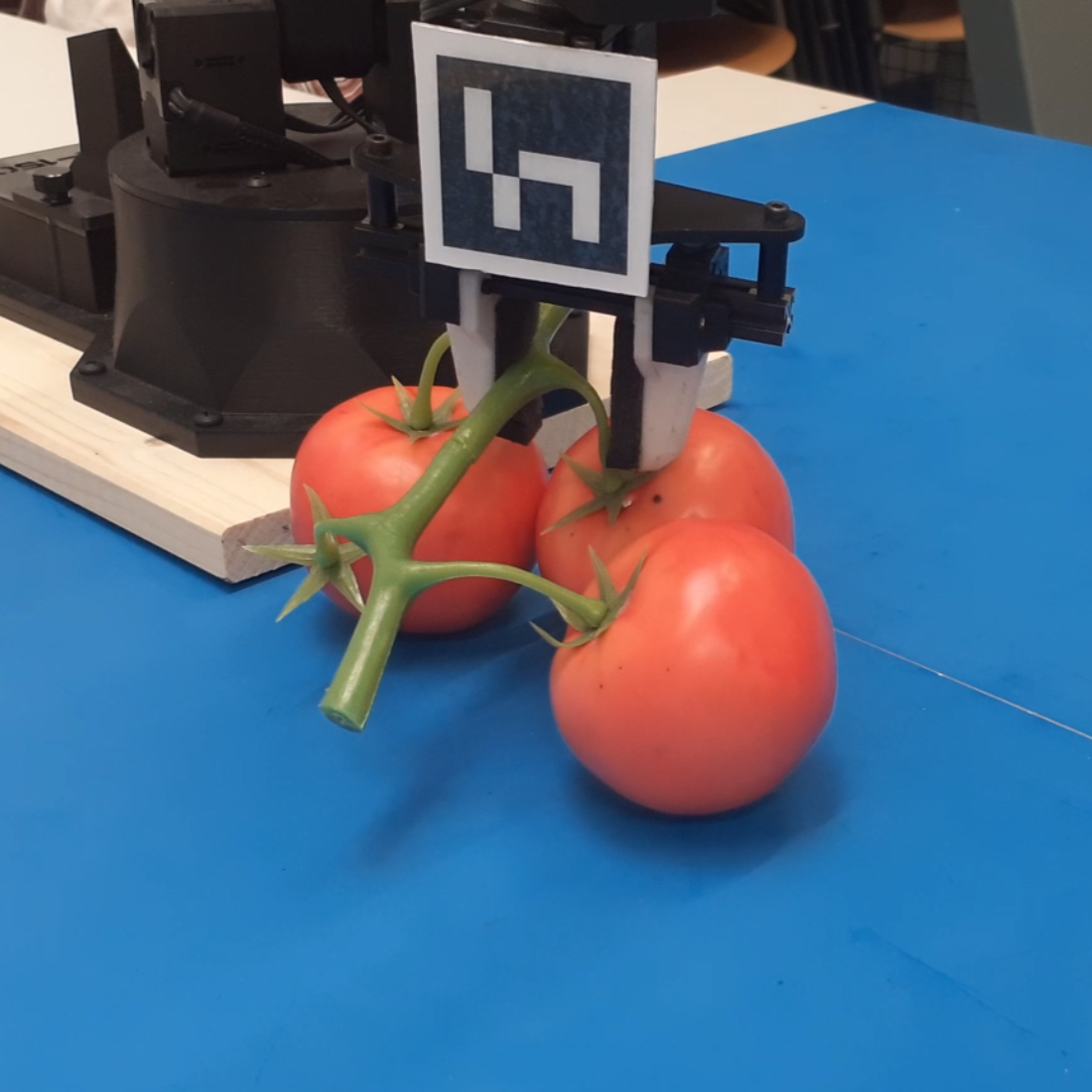}
                }
    \end{minipage}
        \begin{minipage}{\linewidth}
            \centering
            \vskip 0pt
        \subfloat[The model tilts during a grasp. \label{fig:06:result:c}]{
                \includegraphics[height=3.8cm]{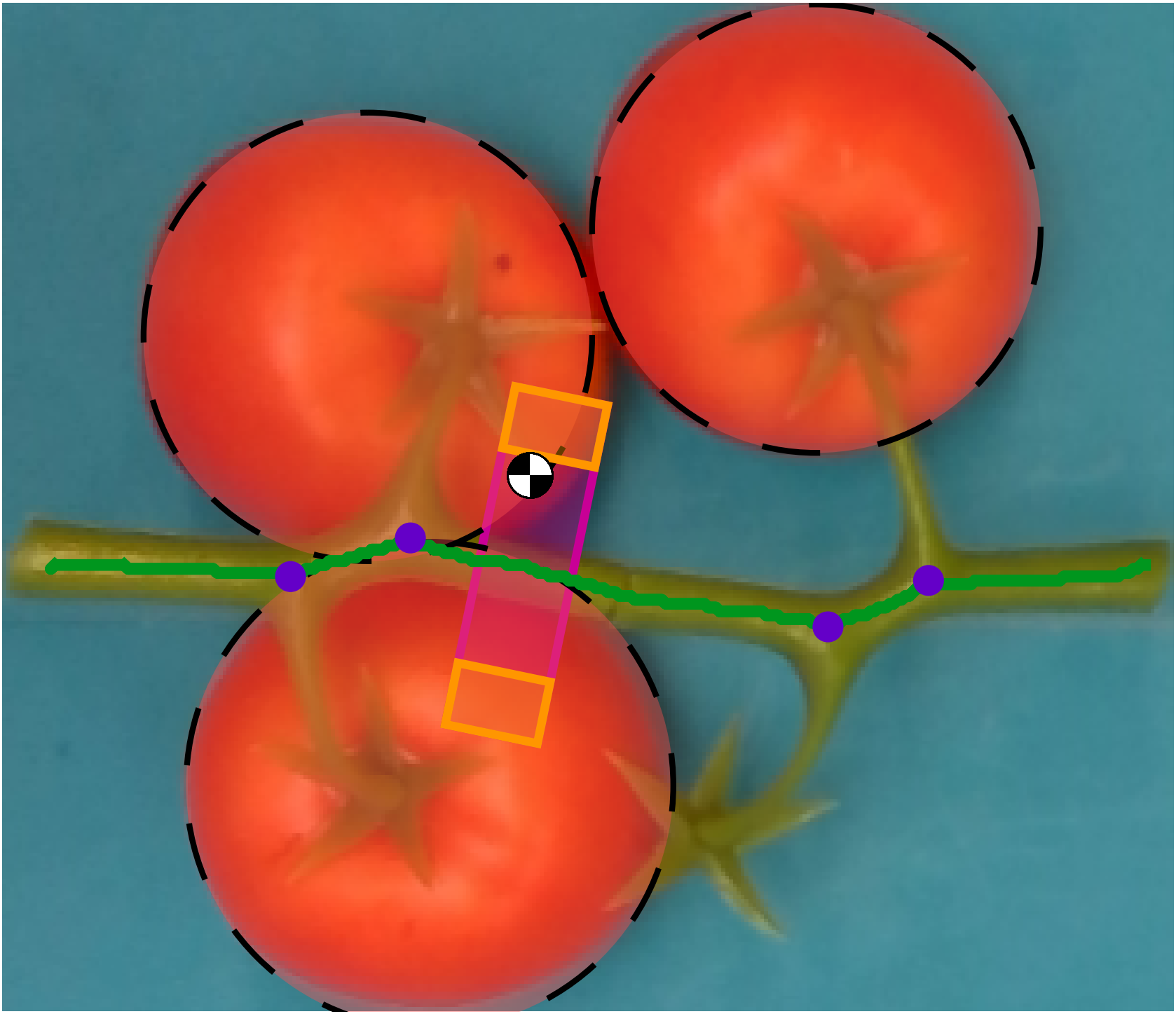}
                \hfill
                \includegraphics[height=3.8cm]{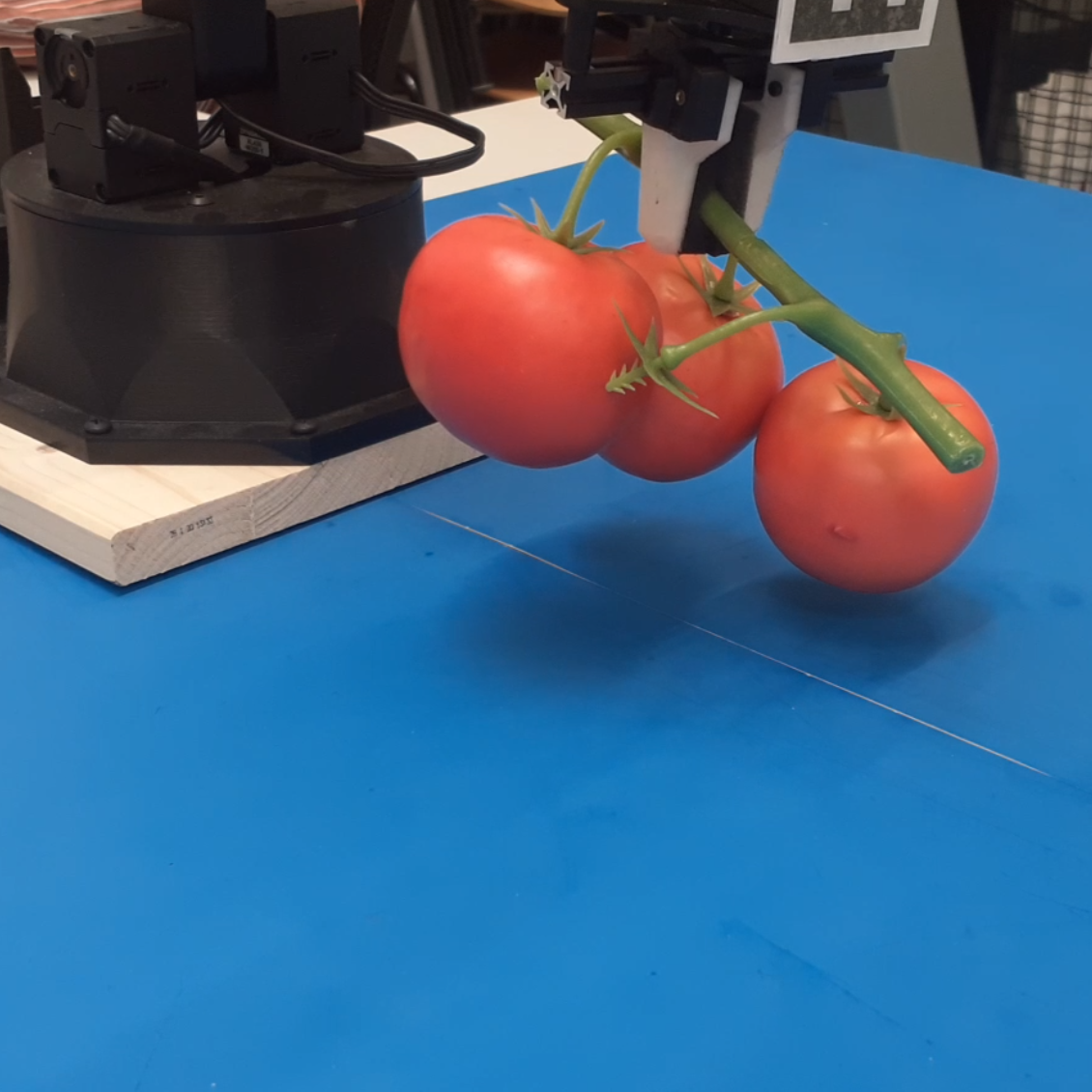}
                }
    \end{minipage}
        \caption{A pick and place routine executed on simple truss model two. The left column shows the planned grasp, and the right column shows the grasp applied to the truss. The tomatoes are represented by the dashed circles, the peduncle by the dark green line, the junctions by the purple dots, the center of mass by the crossed circle, and the opened end effector fingers by the orange boxes, which will be close along the purple lines.}
        \label{fig:06:result}
\end{figure}

The previously described pick and place routine was executed 20 times per truss, such that the truss is grasped from several different poses. The failure rates are reported in Table~\ref{tab:results}, and typical grasps are illustrated in Fig.~\ref{fig:06:result}. Videos taken during the experiment can be found via SurfDrive\footnote{\url{https://surfdrive.surf.nl/files/index.php/s/euEKKhtSRqB9mPG}} and a video attachment is accompanying this paper.
\begin{table}[htbp]
        \vspace{5mm}
        \centering
        \caption{Failure rates of the proposed geometry-based grasping method.}
        \label{tab:results}
        \begin{center}

        \begin{tabular}{m{0.2cm} l m{1cm}  l| l | l l}
                                                        &                       &                       &                       & \multicolumn{3}{c}{failure rate (number of failures)}\\
                                        & \multicolumn{2}{l}{model id}  & attempts        & all   &vision         &
                \begin{minipage}[t]{0.13\columnwidth} planning \& control  %
\end{minipage}  \\ \hline
                \multirow{4}{*}{\rotatebox[origin=c]{90}{simple}}
            & \textbf{all}      &\includegraphics[height = 0.6cm]{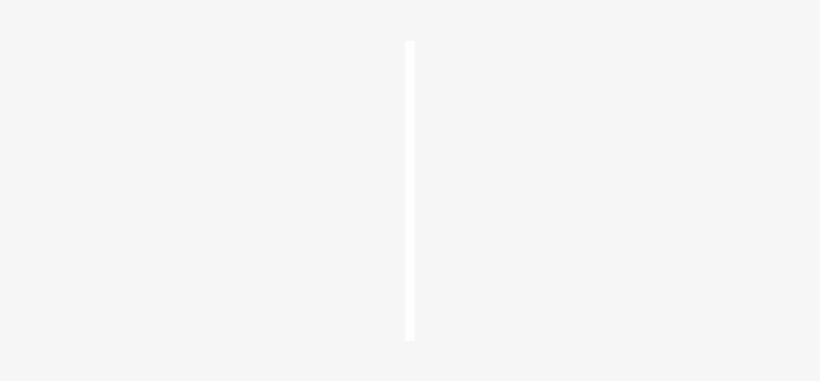} & \textbf{60}           & \textbf{8\% (5)}      & \textbf{3\% (2)}      & \textbf{5\% (3)}        \\[1ex]
                                                        & 1     & \includegraphics[height = 0.6cm]{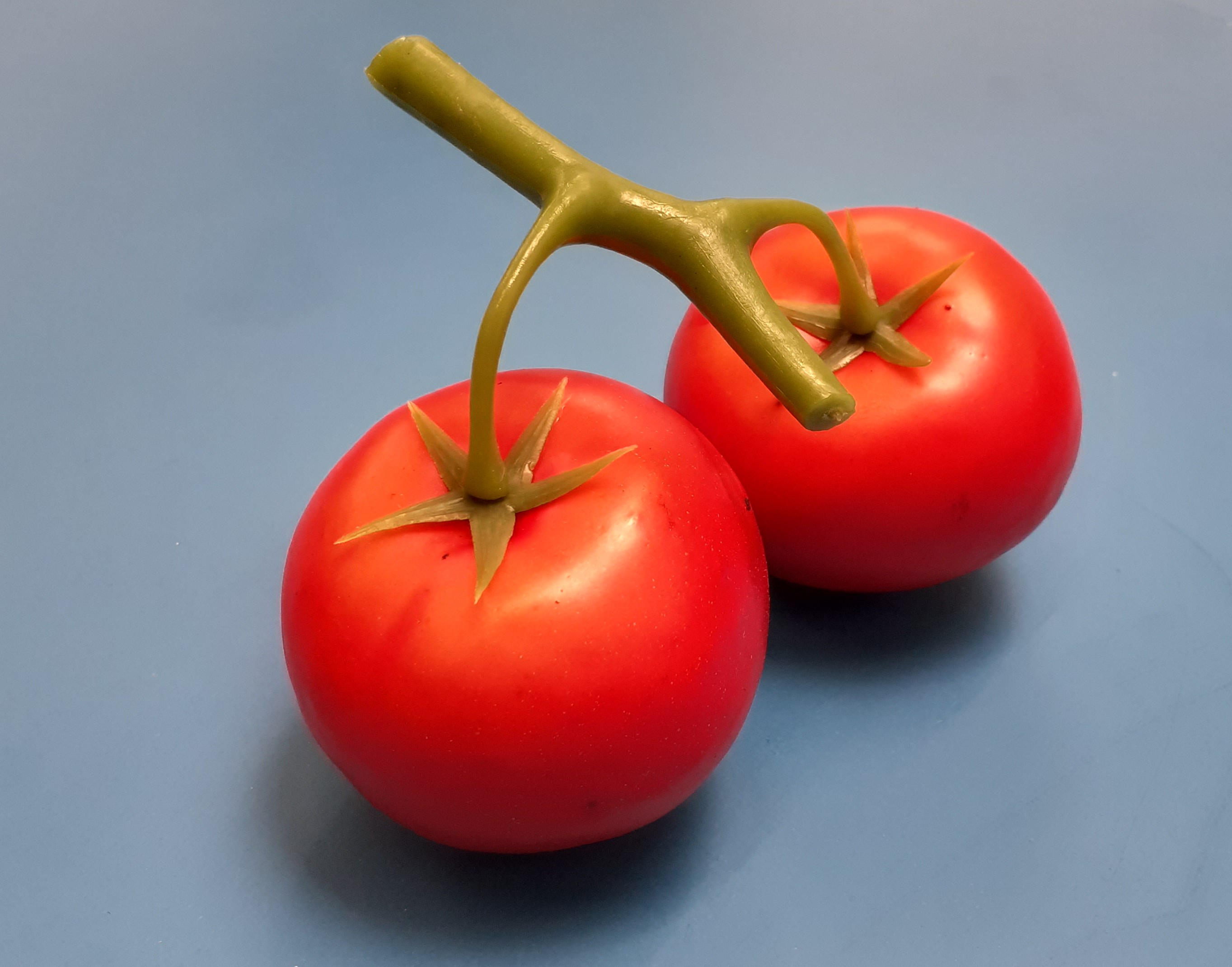}          & 20            & 5\% (1)       & 5\% (1) & 0\% (0)       \\
                                                        & 2     & \includegraphics[height = 0.6cm]{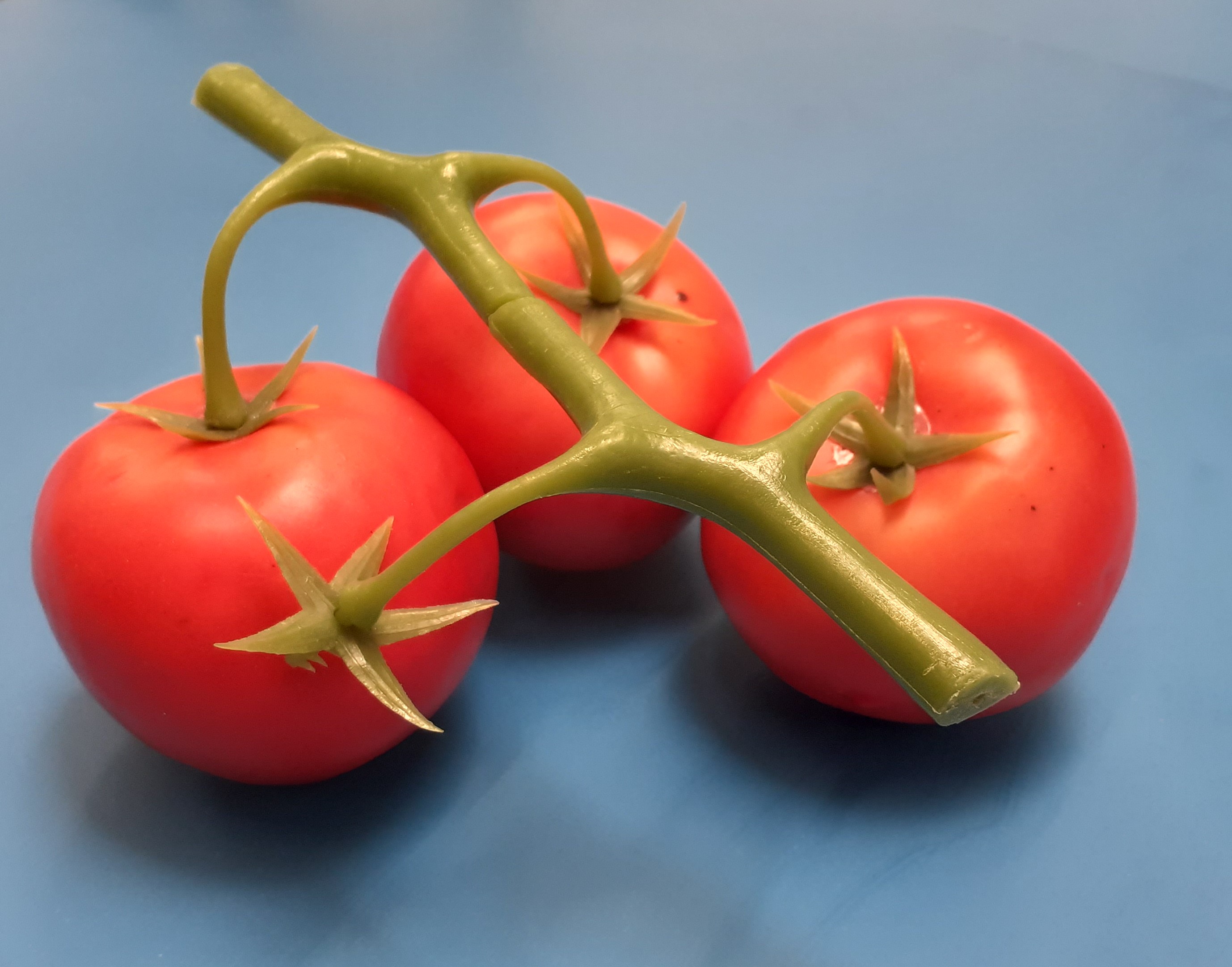}          & 20            & 20\% (4)      & 5\% (1) & 15\% (3)              \\
                                                        & 3     & \includegraphics[height = 0.6cm]{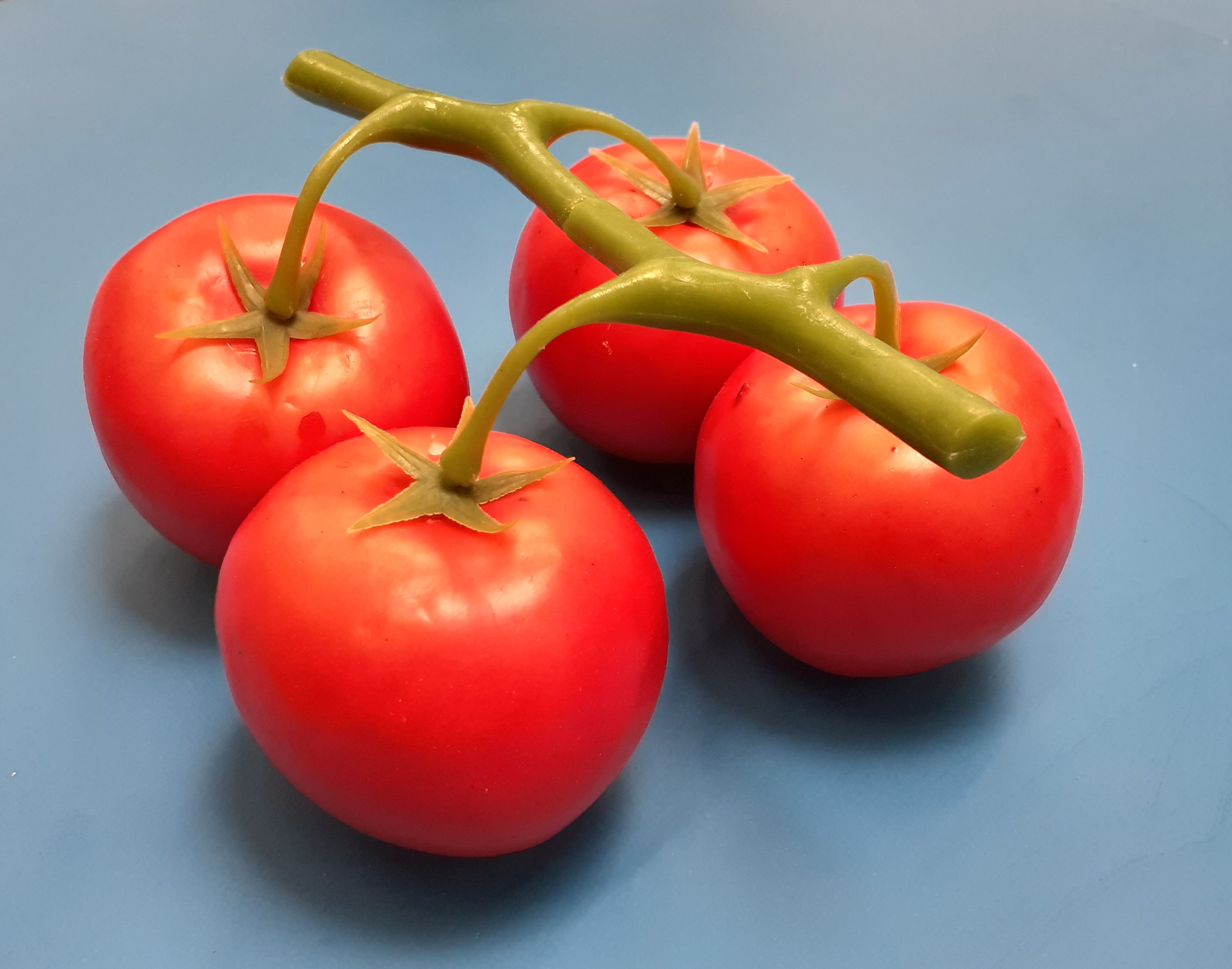}          & 20            & 0\% (0)       & 0\% (0) & 0\% (0)       \\ \hline

                \multirow{4}{*}{\rotatebox[origin=c]{90}{realistic}}  & \textbf{all}    & \includegraphics[height = 0.6cm]{fig/height.png}           & \textbf{60}           & \textbf{17\% (10)}      & \textbf{0\% (0)}      & \textbf{17\% (10)}    \\
                                                        & 1     & \includegraphics[height = 0.6cm]{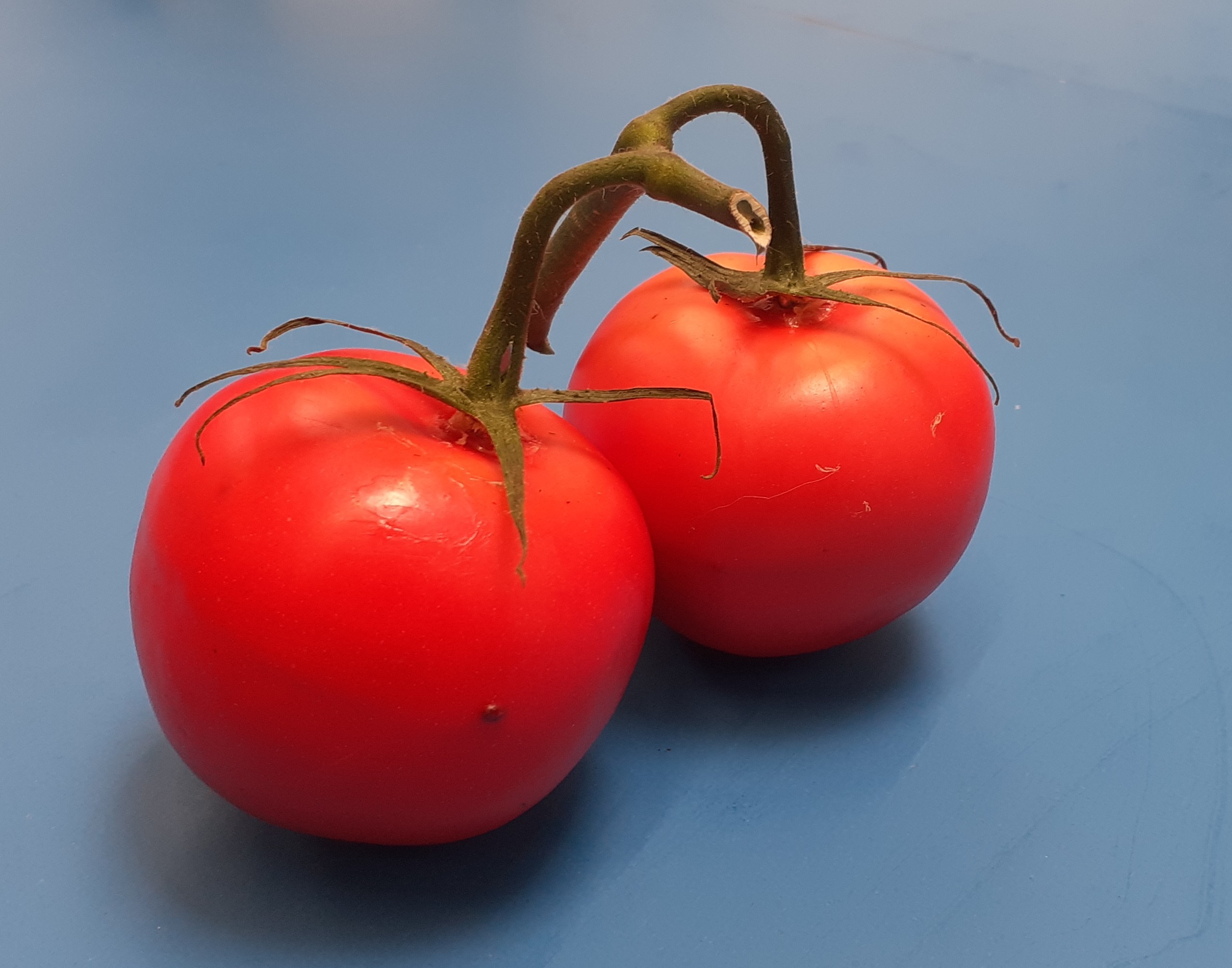}          & 20            & 0\% (0)       & 0\% (0) & 0\% (0)       \\
                                                        & 2     & \includegraphics[height = 0.6cm]{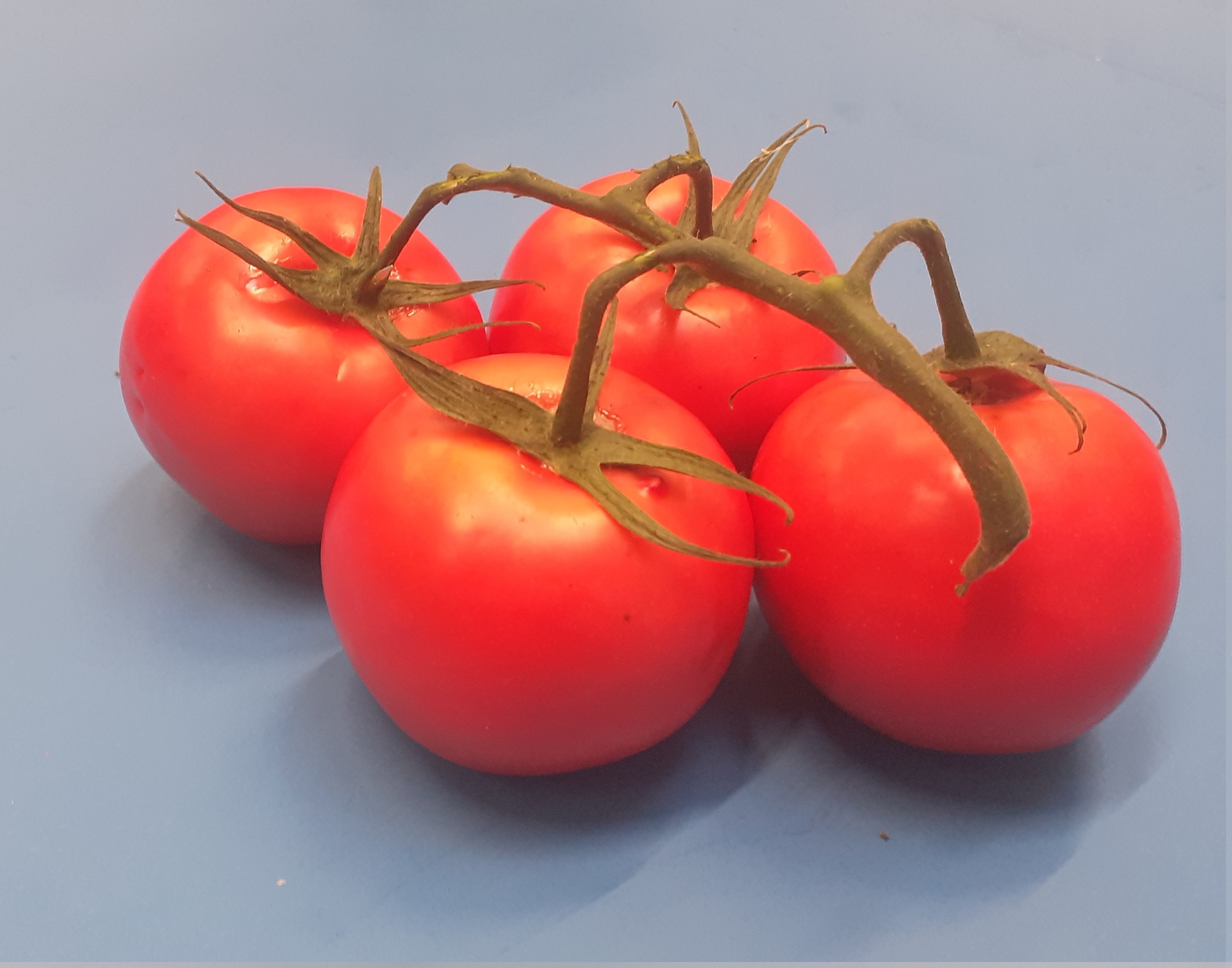}          & 20            &25\% (5)       & 0\% (0) & 25\% (5)      \\
                                                        & 3     & \includegraphics[height = 0.6cm]{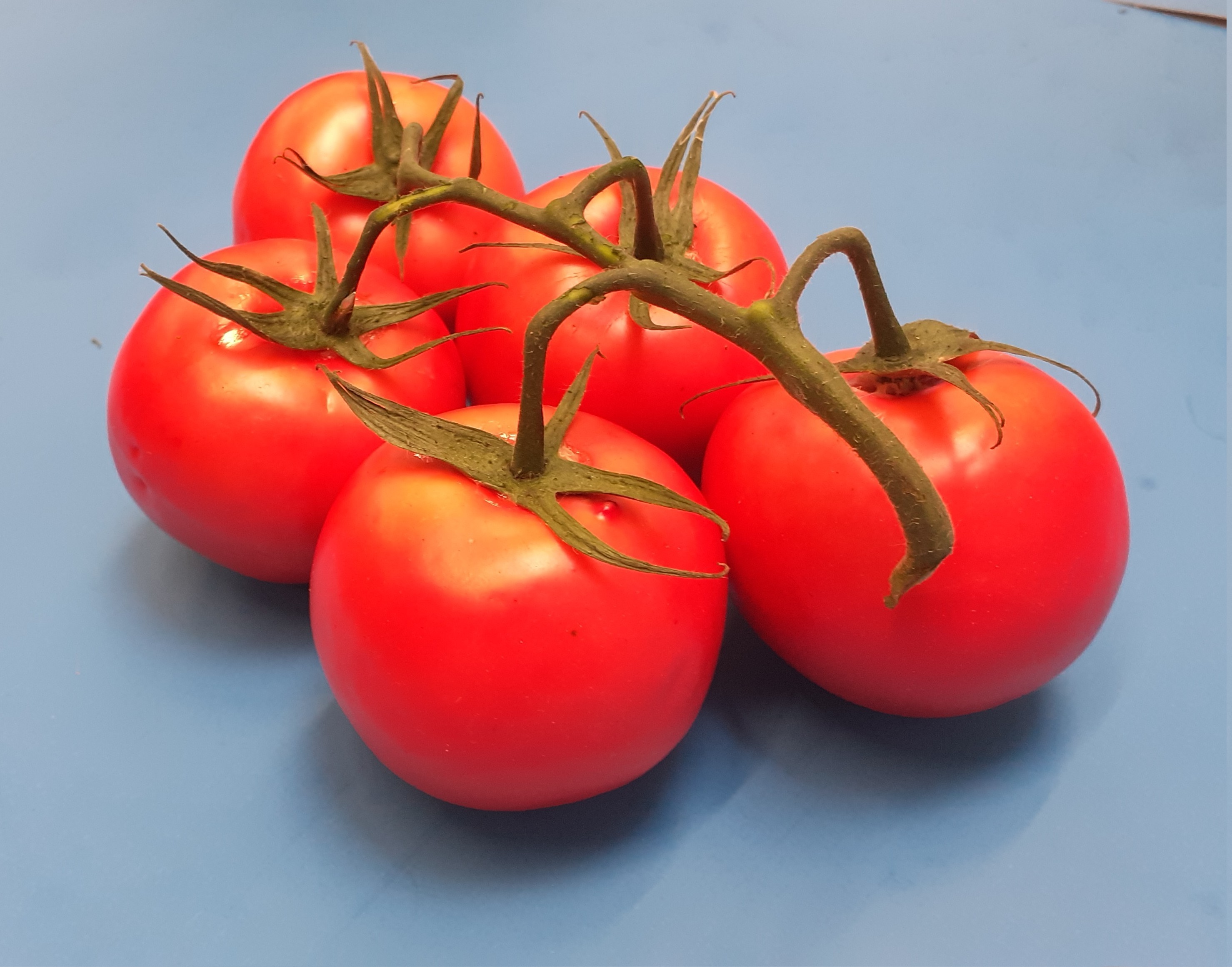}          & 20            &25\% (5)       & 0\% (0) & 25\% (5)      \\
        \end{tabular}
        \end{center}
\end{table}

% simple
A failure rate of 8\% (5/60) was obtained with the simple models. A typical successful grasp is shown in Fig.~\ref{fig:06:result:a}. From the five failures, two were caused by vision, two by planning, and a single failure was caused by control.
For both vision failures, the algorithm was unable to identify the peduncle.
In both planning errors, the algorithm did not take into account the surrounding pedicels, causing the end effector to be blocked. An example is shown in Fig.~\ref{fig:06:result:b}.
The last failure was caused by control, where the end effector closed above the peduncle. Most failures occurred on model two, a single failure was made on model one, and no failures were made on model three. Undesired tilting of the second model was frequently observed as shown in Fig.~\ref{fig:06:result:c}.

 % realistic
On the realistic models, a failure rate of 17\% (10/60) was obtained. Twice as many failures were reported compared to the simple models.
All ten failures were caused by pedicels blocking the end effector. Four of these cases were caused by planning, and six cases were the result of control. No failures were caused by the computer vision system.
All grasp attempts on the first model succeeded, an equal number of failures was observed for models two and three. Most failures on model two were caused by planning, on model three control errors occurred more frequently.

%%%%%%%%%%%%%%%%%%%%%%%%%%
%%%%%%% DISCUSSION %%%%%%%
%%%%%%%%%%%%%%%%%%%%%%%%%%
\section{Discussion}
\label{sec:discussion}
%\todo[inline]{Taeke: Discussion and conclusion still need rewriting, I think I should do this when I know what to tell in the rest of the paper. I also want to go trough some similar paper to see how they discuss and conclude their work.}

In the presented experiments, a significant number of errors were caused by the manipulator control.  However, not all failures can be attributed to the hardware.
For trusses that require more accurate grasping, a significant number of planning errors were made.
On the third simple model the center of mass, and thus also the target grasp pose, is close to a pedicel. At such a location small errors in planning result in grasp failure, which explains the lower success rate for this model. The realistic models have less space around the stem, and the pedicels are located closer to each other. This makes the planning and control of a successful grasp more difficult for these models than the simple models.

The grasp pose is determined based on several criteria, among which the space available at the target location on the peduncle. However, it ignores the surrounding stem parts and tomatoes which may block the grasp. The two-dimensional truss model cannot be used to avoid these surrounding objects, as depth information is vital.
To make a more appropriate trade-off, the point cloud of the tomato truss should be analyzed instead of computing a target grasp location based on a two-dimensional image.
Furthermore, a grasp is always approached in the vertical direction from above as it is assumed that the truss lies on a flat surface. By allowing for different end effector orientations and approach directions, the end effector could grasp the peduncle at previously unreachable locations.
%
% The collision detection and contact determination system from GraspIt! can be used to identify a target grasp location on a given point-cloud \cite{miller2004graspit}. For implementation, the caging-based grasping constraints need to be included and a grasp quality metric based on the amount of free space and distance to the truss center of mass. One of the challenges is to obtain an accurate point cloud of the tomato trusses.

% weakness experimental method
The used experimental method has several limitations.
First of all, only a small number of trusses were used. No reliable conclusions can be drawn about the success rate of the proposed method on vine tomatoes in general.
Secondly, only truss stems with sufficient space around the peduncle to fit the gripper fingers were used. Trusses exist where the peduncle lies much closer to the tomatoes. For such trusses, the proposed geometry-based method will be much more difficult to apply.
Finally, due to hardware limitations, artificial tomatoes were used for the experiments, which may have affected the experimental results. % imposes some simplifications on the pick and place task.
%
\iffalse
First of all, the unrealistic artificial tomato weight may affect the obtained results. The truss stem does not bend as it would under the weight of real tomatoes, and the assumption that the peduncle mass is negligible compared to the tomato mass does not hold for these lightweight tomatoes. As a result, the predicted center of mass is off, causing the truss to tilt.
%  of 10gr can not be neglected compared to the mass
Secondly, the artificial tomatoes encompass much less geometric variety than the real fruit, possibly making the grasping task less complex.
%
Finally, the damage done to these fragile objects could not be measured, which is an essential aspect to automate the packing task of vine tomato.
%
\fi

%%%%%%%%%%%%%%%%%%%%%%%%%%
%%%%%%% CONCLUSION %%%%%%%
%%%%%%%%%%%%%%%%%%%%%%%%%%
\section{Conclusion}
\label{sec:conclusion}

In this paper, we proposed
% 1
a geometry-based grasping method for vine tomatoes. This method relies on geometric information and position-based control. A geometric model of the vine tomato and of the end effector was developed. These models were used to verify constraints for a caging-based grasp and to determine a suitable grasp pose.
% 2
A computer vision pipeline was developed to determine the numerical values of several geometric truss features.
% 3
Finally, the proposed computer vision pipeline and the geometry-based grasping were tested on a robotic setup.
%

% Strengths
The computer vision pipeline is able to obtain the key features for trusses of various shapes and sizes. On the test dataset, 99.7\% of the tomatoes were correctly detected, without reporting any false negatives. Prediction errors on the tomato locations and dimensions resulted in an error of $5.02 \pm 3.26$\,mm on the truss center of mass. The graph-based peduncle detection method identified the peduncle correctly on trusses for which previously reported
methods have failed. The algorithm correctly predicted 86\% of the locations where the stem splits, but also 34\% false positives were reported.
The real-robot experiments show that the newly developed grasping method is capable of grasping vine tomatoes. Experiments were performed on simple trusses with a plastic stem, and on realistic trusses with a real vine tomato stem. On the simple models, 92\% of the pick and place attempts succeeded, while
on the realistic models, 83\% pick and place attempts were successful.

This is a first step in the direction of automating the packaging task of vine tomatoes. Further research is needed to develop extensions required for real-world implementation.
Experiments should be performed with high-end hardware and real tomatoes, such that the damage to the crop can be analyzed.
The presented method does not take into account the surrounding stem parts or tomatoes when computing a target grasp location.
A point cloud of the tomato truss should be used such that these surrounding objects can be avoided.
Other extensions are to conduct experiments on a larger batch of trusses and to deal with multiple touching and overlapping trusses.

%\todo[inline]{Taeke: When to use passive form?}
%\todo[inline]{Taeke: How to refer to PicknPack? Information is scattered over many different reports, and some information is obtained via e-mail}
\balance
\section*{Acknowledgment}
This research is funded by the Netherlands Organization for Scientific Research project Cognitive Robots for Flexible Agro-Food Technology, grant P17-01.
\bibliographystyle{IEEEtran}


\end{document}